\documentclass[sn-nature]{sn-jnl}

\usepackage{graphicx}
\usepackage{textcomp}
\usepackage{multirow}
\usepackage{array}
\usepackage{url}
\usepackage{enumerate}
\usepackage{bm}
\usepackage{amsmath}
\usepackage{amssymb} 
\usepackage{amsfonts}  
\usepackage{amsthm}  
\usepackage{mathrsfs}  
\usepackage[title]{appendix}  
\usepackage{xcolor}  
\usepackage{manyfoot}  
\usepackage{booktabs}
\usepackage{algorithm}
\usepackage{algorithmicx}  
\usepackage{algpseudocode}
\usepackage{listings}  

\theoremstyle{thmstyleone}%
\theoremstyle{thmstyletwo}%

\theoremstyle{thmstylethree}%
\begin{document}

\title[Diffuse-UDA]{Diffuse-UDA: Addressing Unsupervised Domain Adaptation in Medical Image Segmentation with Appearance and Structure Aligned Diffusion Models}


\author[1,2]{\fnm{Haifan} \sur{Gong}}

\author[1,2]{\fnm{Yitao} \sur{Wang}}

\author[1,2]{\fnm{Yihan} \sur{Wang}}

\author[1]{\fnm{Jiashun} \sur{Xiao}}

\author[1,3]{\fnm{Xiang} \sur{Wan}}

\author*[1]{\fnm{Haofeng} \sur{Li}}\email{lhaof@foxmail.com}

\affil[1]{\orgname{Shenzhen Research Institute of Big Data}}

\affil[2]{\orgdiv{SSE}, \orgname{The Chinese University of Hong Kong (Shenzhen)}}

\affil[3]{\orgname{Guangdong Provincial Key Laboratory of Big Data Computing}}










\abstract{The scarcity and complexity of voxel-level annotations in 3D medical imaging present significant challenges, particularly due to the domain gap between labeled datasets from well-resourced centers and unlabeled datasets from less-resourced centers. This disparity affects the fairness of artificial intelligence algorithms in healthcare. We introduce Diffuse-UDA, a novel method leveraging diffusion models to tackle Unsupervised Domain Adaptation (UDA) in medical image segmentation. Diffuse-UDA generates high-quality image-mask pairs with target domain characteristics and various structures, thereby enhancing UDA tasks. Initially, pseudo labels for target domain samples are generated. Subsequently, a specially tailored diffusion model, incorporating deformable augmentations, is trained on image-label or image-pseudo-label pairs from both domains. Finally, source domain labels guide the diffusion model to generate image-label pairs for the target domain. Comprehensive evaluations on several benchmarks demonstrate that Diffuse-UDA outperforms leading UDA and semi-supervised strategies, achieving performance close to or even surpassing the theoretical upper bound of models trained directly on target domain data. Diffuse-UDA offers a pathway to advance the development and deployment of AI systems in medical imaging, addressing disparities between healthcare environments. This approach enables the exploration of innovative AI-driven diagnostic tools, improves outcomes, saves time, and reduces human error.}



\maketitle
\section{Introduction}

\begin{figure}[!tbp]
\includegraphics[width=1\linewidth]{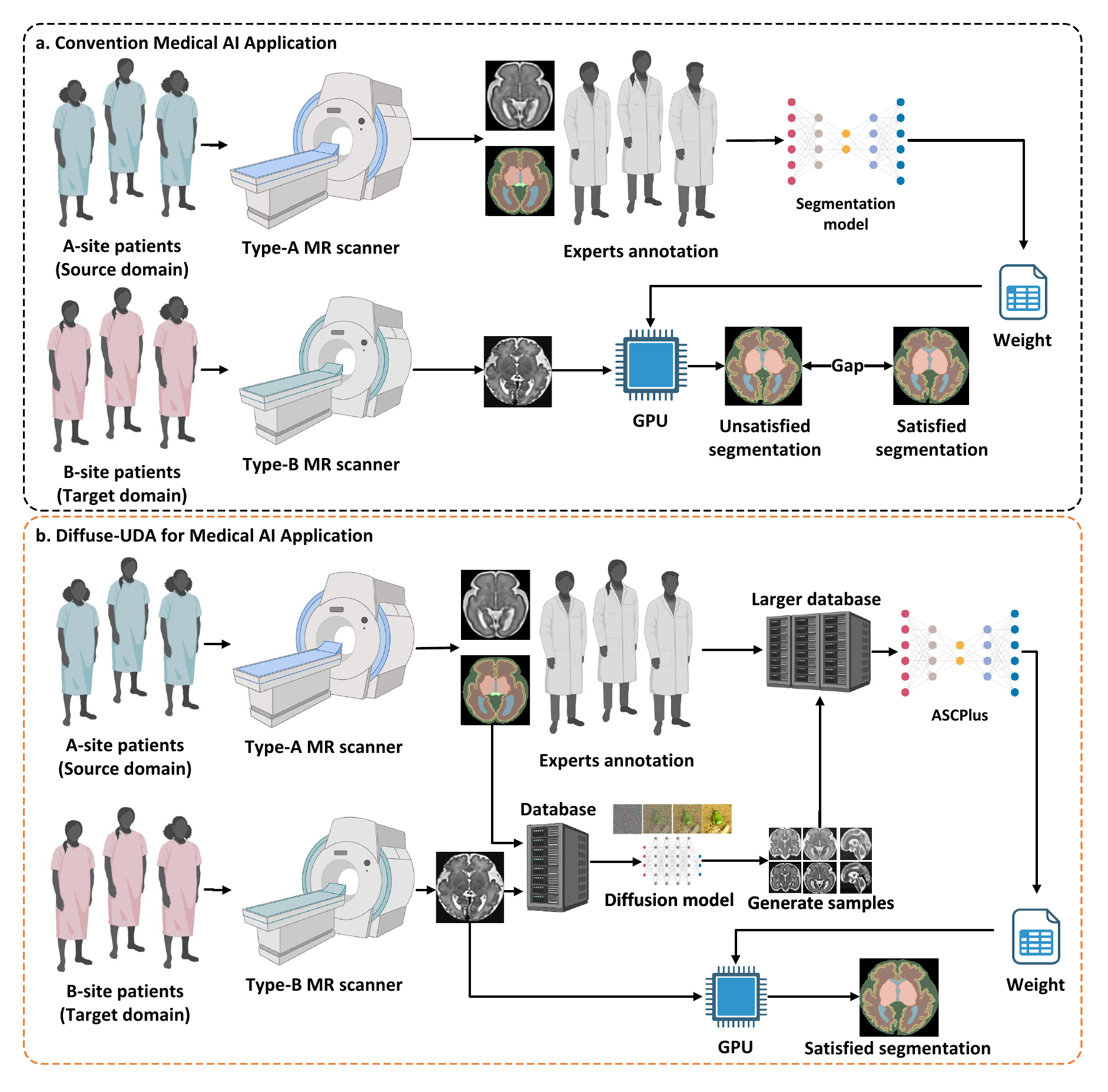}
\centering
\caption{Comparison of conventional medical AI applications and our DiffuseUDA medical AI application. a. Conventional medical AI applications are trained on annotated data from Center A and used at Center A, but their performance often decreases significantly when transferred to Center B due to domain differences in imaging and patient conditions. b. The proposed DiffuseUDA method utilizes a diffusion model to generate new data based on labeled data from Center A and unlabeled data from Center B, with an additional style-structure consistency network to ensure the trained model's usability when switching to other data centers.}
\label{fig:diffuse-uda}
\end{figure}

Recent advances in deep learning (DL) have significantly advanced the field of automatic medical image segmentation~\cite{makropoulos2018review,wang2021annotation,gong2024nnmamba,gong2024intensity}, a critical component in the detection and treatment of illnesses~\cite{isensee2021nnu,gong2023thyroid}. The efficacy of DL-based models, however, is contingent upon both the volume and quality of the available data. This dependency poses a challenge particularly when labeled data is confined to one data center and the model is to be applied to another~\cite{guan2021domain,gao2023synthetic,xu2023asc,ktena2024generative}. Variances in imaging devices and hyper-parameters across data centers give rise to a \emph{domain shift}, a discrepancy wherein a model trained on data from one domain underperforms on data from another. Compounding this issue is the fact that medical images are typically represented as 3D voxels, rendering the labeling process in a new data center exceedingly laborious~\cite{chen2020unsupervised,payette2021automatic}. These typical situations are shown in Fig.\ref{fig:diffuse-uda}.

To mitigate the performance degradation induced by \emph{domain shift}, considerable research efforts have been directed towards the unsupervised domain adaptation (UDA) task within the realm of medical image segmentation~\cite{chen2020unsupervised,al2021olva,han2021deep,xu2023asc,wang2024towards}. UDA represents a formidable yet practical challenge within real-world scenarios. The objective of UDA is to facilitate high-quality predictions on target domain data in the absence of target labels~\cite{xu2019larger}. This task is particularly prevalent in the medical field, where variations in imaging devices and parameters across hospitals necessitate robust adaptation strategies~\cite{xu2023asc}. Current leading approaches to UDA can be categorized into semi-supervised learning-based methods—including teacher-student frameworks~\cite{bai2023bidirectional,wang2023dhc}, pseudo-labeling techniques~\cite{zhao2022semi,wang2024towards}, and contrastive learning paradigms~\cite{zhao2022cross}—as well as feature alignment methods, such as generative adversarial learning~\cite{al2021olva} and data augmentation strategies~\cite{yun2019cutmix,xu2023asc,zhao2023masked}. Our contribution seeks to harness the capabilities of diffusion models to enhance feature representation learning on target domain data. The diffusion model has recently achieved significant success in generating high-quality visual content by capturing the training data's underlying distribution~\cite{ho2020denoising,song2021score,yang2023diffusion}. This success extends to the medical field, with notable applications in image generation~\cite{chung2022score,yu2023diffusion} and segmentation~\cite{wu2023medsegdiff,wu2024medsegdiff}. In our work, we explore the diffusion model's potential to alleviate domain discrepancies across diverse medical datasets and introduce the Diffuse-UDA module as a plug-and-play enhancement to bolster the performance of existing segmentation frameworks.
Despite the adage that ``\textit{Opinions are like noses, everyone has one.}'', a pivotal question persists within UDA research: \emph{what are the essential factors for successful unsupervised domain adaptation?} By answering above question, we proposed a simple \& effective, scalable, plug-and-play framework named Diffuse-UDA, for addressing unsupervised domain adaptation tasks in medical image segmentation, which is shown in the following perspectives:

\noindent \textbf{(i) Importance of Pseudo Labels:} Let's consider an upper-bound situation in which we could label the target domain's training data perfectly. This will ensure the best possible results on the target domain's test data. This ideal situation encourages works~\cite{zhao2022semi,lee2013pseudo,basak2023pseudo,yun2019cutmix,wang2022semi,yao2022enhancing} that focus on creating good pseudo labels in the target domain to boost performance. Recent advancements in diffusion models have significantly improved the synthesis of high-quality, realistic medical images~\cite{ho2020denoising,song2021score,fan2023survey}. Motivated by this progress, we leverage pseudo labels to guide the diffusion model in generating high-fidelity image-label pairs. In our framework, the diffusion model serves as a robust generator, utilizing images from both the source and target domains as reconstruction objectives, conditioned on corresponding labels and pseudo labels. This approach effectively reduces biases associated with unreliable pseudo-labels. Moreover, we use the model ensemble strategy in the inference process of our backbone UDA model, making it able to get more precise pseudo labels.

\noindent \textbf{(ii) Impact of Appearance and Structure:} In UDA, it's difficult to get pseudo labels for the target domain that are just as good as labels made by experts. The reason is the gap in structure and appearance between the source and target domains, and a typical illustration of these gap are shown in Fig.~\ref{fig:situations}. This issue has led to studies~\cite{xu2023real,xu2023asc,zhao2023masked,de2022adverse,wang2024towards} that try to mitigate these differences. In addition to its robust generative capabilities, Diffuse-UDA facilitates the alignment of visual characteristics between the source and target domains due to the reconstruction of images influenced by both domains. To accommodate the diverse structures present in the target domain, we have incorporated a straightforward deformation augmentation module. This module simultaneously augments the imaging data and its associated masks, thereby enhancing the model's generalizability to previously unseen data.

\noindent \textbf{(iii) Data Is Always Needed:} If we use more training data, the results get better. The rule still exists for the UDA task and other transfer learning tasks~\cite{he2022masked,li2023well,rastogi2021data}. Regarding the training data scale, our Diffuse-UDA approach is inherently scalable with the deformable augmentation module for sampling, permitting the straightforward expansion of the training dataset, which consequently leads to improved model performance.

Our methodology, termed Diffuse-UDA, strategically circumvents common pitfalls associated with pseudo-label inaccuracies through a multi-stage approach. In the initial phase, our proprietary ASCPlus algorithm is employed to significantly enhance the accuracy of pseudo-label generation for target domain data. This foundational step ensures the subsequent stages operate on a more reliable dataset, setting a solid groundwork for effective domain adaptation.
In the second stage training process, we integrate both pristine source domain label-image pairs and the refined target domain label-image pairs into our training regimen. This dual-input strategy not only utilizes the strengths of each domain but also actively diminishes the adverse effects typically caused by unreliable pseudo-labels. By merging these diverse data sources, we effectively enrich the training process, leveraging the distinct characteristics inherent in each domain.
Moreover, the second stage inference process includes a sophisticated sampling technique that allows us to scale up our training dataset substantially. Due to the varied origins of these data points, our approach inherently ensures a diverse array of generated data styles. We further enhance this diversity by applying deformable transformations to the sampling conditions, specifically the labels, which introduces a broader spectrum of anatomical structures into our training set. This variety is crucial for the model's ability to generalize across different imaging conditions and anatomical variations.
These methodological enhancements are instrumental in achieving optimal segmentation performance in the final stage of our process. By systematically addressing the challenges at each stage of model training, Diffuse-UDA sets a new standard for segmentation accuracy and model robustness in the field of medical image analysis.

\section{Results}
\subsection{Overview of DiffuseUDA Framework}
We design the Diffuse-UDA, a powerful framework to address the unsupervised medical image segmentation, which is shown in Fig.\ref{fig:pipeline}. As depicted in Fig.\ref{fig:pipeline}-a, our approach, Diffuse-UDA, encompasses three pivotal steps, each with its respective time allocation. Fig.\ref{fig:pipeline}-b explicates the significance of the letters within the accompanying table. Initially, the ASCPlus model is deployed to assign pseudo-labels to unlabeled data from the target domain (Fig.\ref{fig:pipeline}-c). Subsequently, the Diffusion model is utilized, leveraging paired image-mask data to train a conditional diffusion model, thereby generating samples exhibiting diverse appearances and structures using masks sourced from the source domain (Fig.\ref{fig:pipeline}-d). This step is followed by training the ASCPlus model using source domain data and pseudo-labeled data as new source domain data, while the unlabeled data serves as target domain data, ultimately preparing the model for deployment at Center B for prediction purposes (Fig.\ref{fig:pipeline}-e). The architectural design of the ASCPlus network features an exponential moving average (EMA) mechanism, where the student model acquires knowledge from source data $D_{s}$ and frequency-transformed source data $D_{sft}$ through the supervised loss $L_{seg}$, ensuring appearance and structure consistency via the loss $L_{asc}$ (Fig.\ref{fig:pipeline}-f). Lastly, the intricate design of our conditional diffusion model is illustrated, incorporating a deformable augmentation module for robust and varied generation, with a fully connected (FC) layer embedding the condition (label) (Fig.\ref{fig:pipeline}-g).
\subsection{Dataset characteristics for unsupervised domain adaptation}
\begin{figure}[!tbp]
\includegraphics[width=0.95\linewidth]{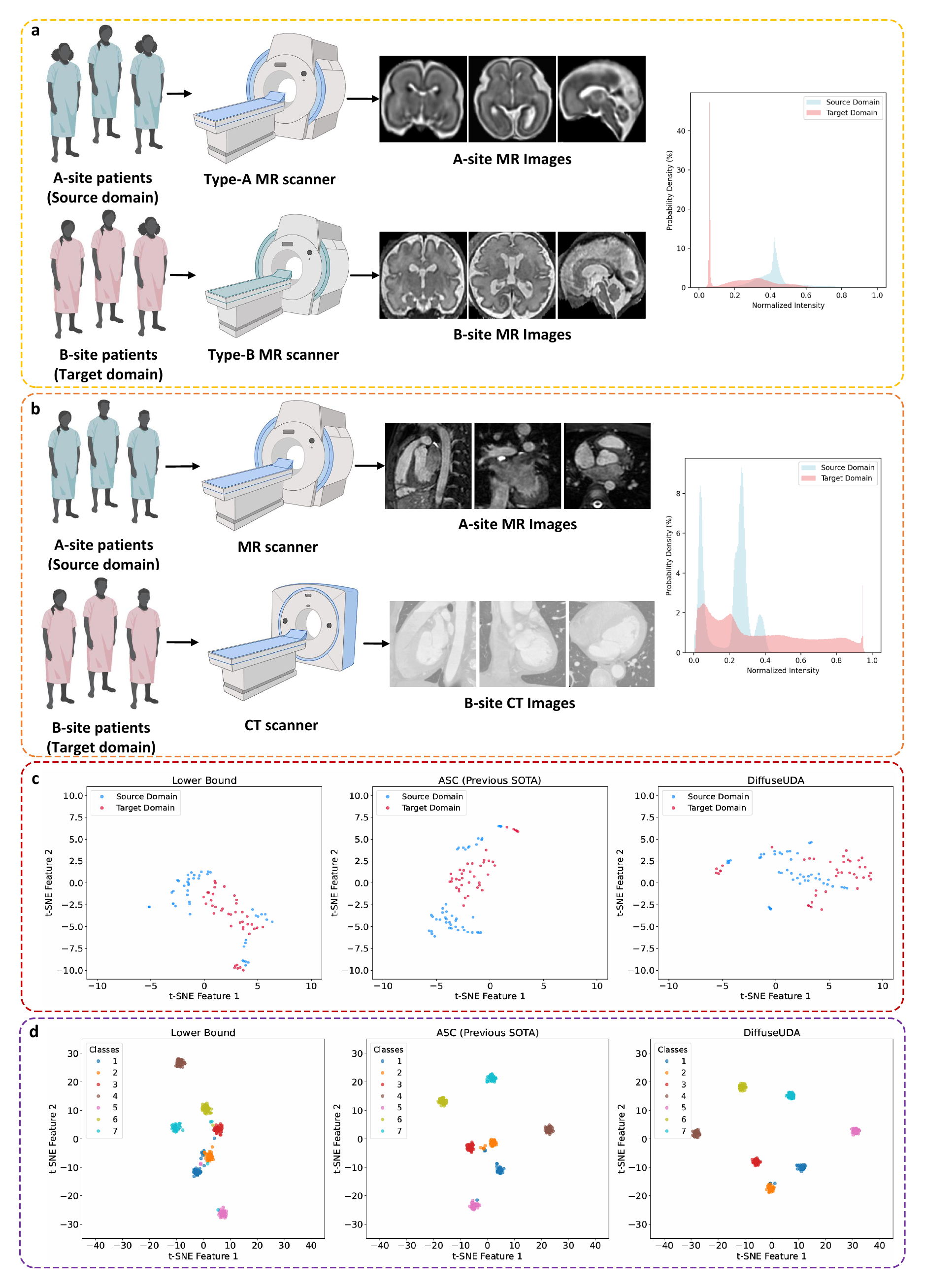}
\centering
\caption{Two typical domain gap situations. a. Patients at Center A use an MR scanner of model A, while patients at Center B use an MR scanner of model B, resulting in differences in individual patient characteristics and imaging equipment parameters. b. Patients at Center A use an MR scanner, while patients at Center B use a CT scanner, leading to differences in individual patient characteristics and imaging modalities. c. Visualization of high-level (feature between encoder and decoder) feature from different domain data. d. Visualization of decoder's ending feature from different domain. }
\label{fig:situations}
\end{figure}

\begin{figure}[!tbp]
\includegraphics[width=1\linewidth]{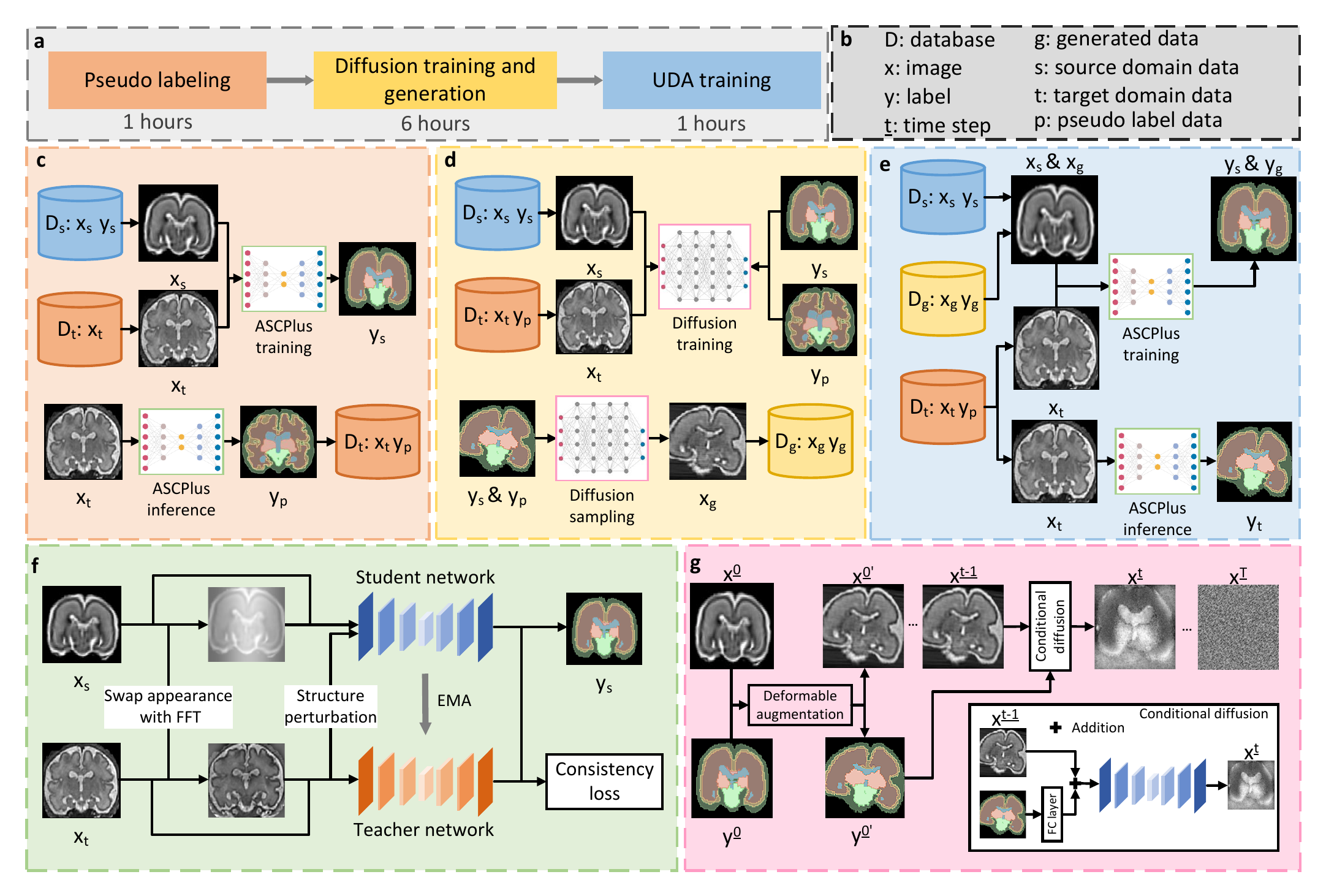}
\centering
\caption{Overview of Diffuse-UDA. a. The three key steps in our process and the respective time required for each; b. The meaning of the letters in this table. c. The first step, using the proposed ASCPlus model, applies pseudo-labels to the unlabeled data from the target domain; d. The second step, using the proposed Diffusion model and the paired image-mask data to train the conditional diffusion model and generate samples with various appearances and structures using masks from the source domain. e. Using the source domain data and pseudo-labeled data as new source domain data, and unlabeled data as target domain data, we train the ASCPlus model for deployment at Center B to provide predictions. f. The proposed ASCPlus network architecture, where EMA represents exponential moving average updates. The student model learns from source data $D_{s}$ and frequency-based transformed source data $D_{sft}$ via the supervised loss $L_{seg}$. The appearance and structure consistency is achieved by the loss $L_{asc}$. g. Detailed design of our conditional diffusion model, which contains a deformable augmentation module for diverse and robust generation. FC layer indicates a fully connected layer for embedding the condition (label).}
\label{fig:pipeline}
\end{figure}

The first dataset focus on addressing the domain gap cross different data center and different scanner type, which is shown in Fig.~\ref{fig:situations} (a). We follow the setting of~\cite{xu2023asc} that take the fetal atlas~\cite{gholipour2017normative,wu2021age,fidon2022spatio} as the source domain and the data in FeTA benchmark~\cite{payette2021automatic} as the target domain. There are T2-weighted MRI volumes with meticulously annotated segmentations for various brain tissues including external cerebrospinal fluid (eCSF), grey matter (GM), white matter (WM), lateral ventricles (LV), cerebellum (CBM), deep grey matter (dGM), and brainstem (BS). The dataset is stratified into two distinct subgroups: 31 scans from neurologically typical fetuses and 49 scans from fetuses exhibiting abnormal developmental patterns. For our source dataset, we utilized a composite of three atlas collections: 32 scans from neurotypical fetal brain atlases~\cite{gholipour2017normative,wu2021age} and 15 scans from atlases of fetuses with spina bifida~\cite{fidon2022spatio}. Comprehensive segmentation annotations for all tissue types were available across the entire atlas collection. In alignment with standard UDA protocols~\cite{huo2018synseg}, we divided the target dataset into two equal subsets, designating 40 volumes for training purposes and the remaining 40 for validation. Specifically, the dataset details are as follows: for the fetal brain segmentation task using T2 MR modality, there are 47 volumes in the source domain for training, and in the target domain, there are 40 volumes for training and another 40 volumes for testing.

The second dataset focuses on addressing the domain gap across different data centers and modalities, which is shown in Fig.~\ref{fig:situations} (b). We use the widely used Multi-Modality Whole Heart Segmentation (MMWHS) challenge 2017 dataset for cardiac tissue segmentation~\cite{zhuang2016multi}. This dataset includes 20 MRI and 20 CT volumes, and their corresponding segmentation ground truth. The image size is either 256×256 or 512×512, and the pixel size varies from 0.28 mm × 0.28 mm to 1.2 mm × 1.2 mm. The number of slices ranges from 112 to 363 with slice thickness varying from 0.45 mm to 1.6 mm. In this dataset, the annotations of four tissues, including ascending aorta (AA), left atrium blood cavity (LAC), left ventricle blood cavity (LVC), and myocardium of the left ventricle (MYO) were provided by the dataset organizer. Specifically, for the MR (source domain) to CT (target domain) modality adaptation task (mr2ct), the dataset is divided into 16 MR slices with its corresponding label for training, 13 CT slices for training without label, 3 CT slices for validation, and 4 CT slices for testing. For the CT (source domain) to MR (target domain) modality transfer task (ct2mr), the dataset is similarly divided into 16 CT slices with its corresponding label for training, 13 MR slices for training without label, 3 MR slices for validation, and 4 MR slices for testing.

\subsection{Unsupervised domain adaptation cross different scanner}
The results presented in Table \ref{exp:sota} compare the performance of our proposed Diffuse-UDA method with various state-of-the-art unsupervised domain adaptation (UDA) and semi-supervised learning (SSL) techniques on the fetal brain dataset, categorized into abnormal and normal fetal brain segments. The upper bound (UB) method, which supervises with $D_{t}$, achieves the highest Dice Similarity Coefficient (DSC) and Normalized Surface Dice (NSD) scores across both abnormal and normal fetal brain categories, indicating the optimal performance when the target domain labels are available. The lower bound (LB) method, supervising with $D_{s}$, shows significantly lower performance, highlighting the challenge of domain adaptation without target domain labels. Among the UDA methods, FDA~\cite{yang2020fda} shows superior performance compared to OLVA~\cite{al2021olva} and DSA~\cite{han2021deep}, yet still falls short of the UB scores. Similarly, within SSL methods, Cutmix~\cite{yun2019cutmix} and PL~\cite{yan2022unsupervised} demonstrate competitive results but do not surpass our proposed Diffuse-UDA.

Our proposed Diffuse-UDA method outperforms all the compared UDA and SSL methods. Some of the results even exceeds the upper bound performance, indicating the effectiveness of our method in bridging the domain gap. The absence of $p$-values for Diffuse-UDA highlights its statistically significant improvement over other methods. The substantial performance gains of Diffuse-UDA emphasize the benefits of incorporating deformable augmentation and scalable mask sampling in the domain adaptation process, validating its robustness and applicability in clinical settings for fetal brain analysis. Visual comparisons presented in Figure~\ref{fig:qualitative} demonstrate that our method yields improved segmentation, particularly in the complex junction areas of brain tissues.

\subsection{Unsupervised Domain Adaptation across Different Imaging Modalities}
The table presents a detailed comparison of state-of-the-art methods in the domain of medical image translation, specifically focusing on the MMWHS benchmark~\cite{zhuang2019evaluation}, which involves the transformation from CT (Computed Tomography) to MR (Magnetic Resonance) imaging and vice versa. The methods are categorized into two main groups: UDA (Unsupervised Domain Adaptation) and SSL (Semi-Supervised Learning), with additional comparisons to established upper (UB) and lower bounds (LB) for performance metrics. In the analysis, each method is evaluated based on two key performance indicators: DSC (Dice Similarity Coefficient) and ASD (Average Surface Distance) following~\cite{wang2024towards}, with results provided for both transformation directions (CT to MR and MR to CT). The table highlights the best-performing methods in bold, emphasizing the advancements in cross-modality medical image segmentation tasks. Notably, the method `Diffuse-UDA` achieves the highest scores, suggesting its effectiveness in handling domain shifts in medical imaging.

Although previous methods utilizing diffusion~\cite{wang2024towards} or GAN-based approaches~\cite{zhu2017unpaired} have been explored, their performance has not been satisfactory. We hypothesize that this is due to their lack of a multi-stage training strategy, which leads to the use of erroneous pseudo-labels—particularly during the initial stages of training—to generate corresponding images. These unclear pseudo-label-image pairs could misguide the model optimization in the wrong direction. Our method, Diffuse-UDA, avoids this issue primarily by employing our ASCPlus algorithm in the first stage, which enhances the generation of more accurate pseudo-labels for the target domain data. In the second stage, we employ both clean source domain label-image pairs and target domain label-image pairs for joint model training, further mitigating the negative impact of unreliable pseudo-labels. Furthermore, our method shows a smaller standard deviation in various experiments, indicating that by generating more pseudo-label samples featuring characteristics of both modalities, our algorithm's training is more stable and reliable.

\subsection{Enhanced Feature Extraction Across Domains by DiffuseUDA}
In Figures 2-c and 2-d, we further demonstrate the capability of DiffuseUDA to effectively extract features across different domains. Feature reduction is implemented via t-SNE, utilizing data from the FeTA UDA dataset. Figure 2-c illustrates the results of high-dimensional feature distributions extracted from both the target and source domains. These features were obtained using the encoder from SegResNet, followed by global average pooling to represent each sample's high-dimensional characteristics. Our results show that, in comparison to previous methods such as the lower bounds and the ASC technique, DiffuseUDA significantly narrows the gap between the domains, as evidenced by the intertwined blue and red clusters in the third panel.

Figure 2-d displays the category-specific features extracted by these methods. We randomly sampled ten feature points from each category in the target domain's cases at the final layer of the decoder. It is evident that our method, compared to the earlier lower bounds and ASC, more effectively distinguishes between different categories—demonstrated by the greater distances between clusters—and exhibits the fewest classification errors, with minimal color mixing among clusters. These results further validate the superiority of our approach in feature extraction across different domains.

\subsection{DiffuseUDA as a Source of Synthetic Training Data}
Beyond the advantages outlined previously, our DiffuseUDA methodology is capable of generating additional training data during its second phase. This synthetic data, which does not include real patient information, exhibits characteristics of both shared domains. This unique feature allows us to enhance the performance of analytical models in new, unrelated domains without direct exposure to domain-specific data. 
Within the scope of this paper, we utilize these synthetic datasets in conjunction with our ASCPlus model to achieve performance levels equivalent to those obtained using annotated data from the target domains, albeit without any direct exposure to target domain labels. This achievement underscores the potential of synthetic data to serve as a robust training alternative in scenarios where annotated data is scarce or privacy concerns preclude its use.
Moreover, the synthetic data generated by DiffuseUDA can be particularly useful for model training at other healthcare data centers. As a practical application of this capability, in this paper, we disclose approximately 500 synthetic datasets for fetal brain segmentation and about 300 datasets for cardiac segmentation. We are confident that these datasets will significantly advance the development of cross-center medical AI applications by providing a diverse and extensive pool of data for training purposes, thereby fostering enhanced model generalization and robustness across different medical imaging tasks.

\section{Discussion and Conclusion}
\subsection{Analysis on diffusion model training and generation}
For a fair comparison, we ablate our Diffuse-UDA using the number of image-label pairs fixed to 40 in all experiments. In the first part, we analyze the effect of different types of data ratio (data from the source domain and target domain ), deformable augmentation, and the effect of the quality from the source domain. We only use the label of the samples from the source domain as the condition to sample the image from the diffusion model.

\subsection{Impact of each module in our Diffuse-UDA}
Based on the above analysis, we further perform the ablation study that uses 87 samples to train the diffusion model and conduct the analysis, with 47 samples from the source domain and 40 samples from the target domain training set. We present the results of the ablation study in Table.~\ref{tab:abla-uda}. The M1 denotes the method only uses the source domain data to train the conditional diffusion model. The M2 denotes using the source domain image-label data and target domain image-pseudo-label data to train the conditional generation model. The M3 denotes the method of adding our deformable transformation module to the framework for generation. The results have shown that our diffusion model can use the target domain pseudo label to boost the generation performance. Moreover, it has also shown that our deformation module is useful. M4 denotes the method that scales up the number of training data by 2 times. The ablation study of the components in our proposed Diffuse-UDA generation on the feta dataset with the pseudo label setting is presented in Table \ref{tab:abla-uda}. ``Source mask'' and ``Target mask'' indicate the use of source domain segmentation and target domain pseudo label segmentation for sampling the image-mask pair, respectively. ``Deformable'' indicates applying deformable augmentation to the masks, while ``Scale-up'' indicates sampling the mask an additional time. Models M1 to M5 utilize only half of the samples in both the source and target domain training sets, whereas M6 employs all samples from both domains and samples the mask four times. The results demonstrate a progressive improvement in DSC from M1 to M6, indicating the efficacy of the proposed components, with M6 achieving the highest DSC of $81.1_{\pm0.2}$.

The ablation study of ours ASCPlus is shown in Table \ref{tab:abla-asc}, and all the results boost our method's performance. ``M1'' represents the lower bound that only trains on the source domain data $D_{s}$. ``M2'' uses the aligned source images $x_{sft}$ for training. The following component are based on $L_{asc}$, which is decoupled as $L^{app}_{con(x_{t})}$, $L^{app}_{con(x_{tfs})}$ and $L^{str}_{con}$. ``M3'' denotes the appearance consistency loss $L^{app}_{con(x_{t})}$ to align distribution from source to target. ``M4'' indicates the dual-view appearance consistency loss to constrain semantic invariance. ``M5'' denotes the structure consistency $L^{str}_{con}$. The ablation study of the components in our proposed Diffuse-UDA generation on the feta dataset with the pseudo label setting is presented in Table \ref{tab:abla-uda}. ``Source mask'' and ``Target mask'' indicate the use of source domain segmentation and target domain pseudo label segmentation for sampling the image-mask pair, respectively. ``Deformable'' indicates the application of deformable augmentation to the masks, while ``Scale-up'' indicates sampling the mask with an additional one time. Models M1 to M5 utilize only half of the samples in both the source and target domain training sets, whereas M6 employs all samples from both domains and samples the mask four times. The results demonstrate a progressive improvement in Dice Similarity Coefficient (DSC) from M1 to M6, indicating the efficacy of the proposed components, with M6 achieving the highest DSC of $81.1_{\pm0.2}$.

\begin{figure}[!tbp]
\includegraphics[width=1\linewidth]{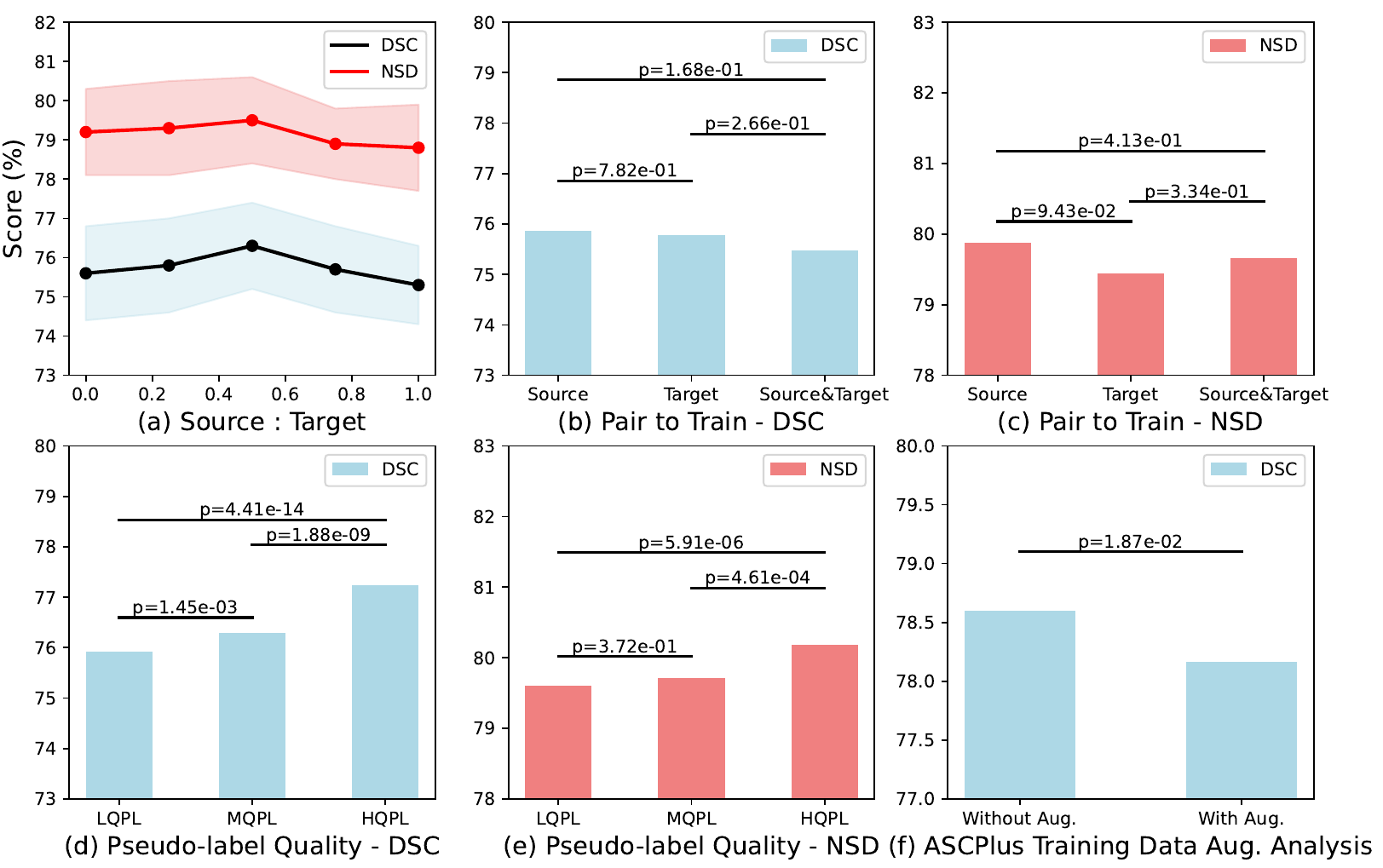}
\centering
\caption{Detailed analysis of using different types of training data construction in our diffuse UDA framework. (a) investigates the effect of using different amounts of sampled data from the source domain and target domain. (b) and (c) indicates the use of the deformation module on different types of training data. (d) and (e) analysis of the effect of different quality of target domain pseudo label (LQPL: low-quality pseudo-label predicted by the PL~\cite{yan2022unsupervised}. MQPL: middle-quality pseudo-label predicted by ASC~\cite{xu2023asc}. HQPL: high-quality pseudo-label predicted by ours ASCPlus.) (f) details the analysis of training ASCPlus with and without our proposed deformable data augumentation module.}
\label{fig:abla-train}
\end{figure}

\noindent \textbf{Analysis on training data for Diffuse-UDA}
Figure~\ref{fig:abla-train} explores various strategies to enhance the performance of our diffusion model under UDA settings. 

\textbf{Figure~\ref{fig:abla-train} (a)} examines the effect of varying the ratio of image-label samples between the source and target domains during training. Observations indicate that even training solely with source domain data enhances performance beyond the baseline method. Optimal results are noted when the data ratio is adjusted to 1:3 (10 from the source domain, 30 from the target domain), 1:1 (20 from each domain), and 3:1 (30 from the source domain, 10 from the target domain), showing significant performance improvements. Conversely, training exclusively with pseudo-labeled target domain images does not yield performance gains, highlighting the importance of the source domain's reliable mapping relationship. This suggests that our diffusion model can effectively generate robust training data and mitigate domain shifts.

\textbf{Figure~\ref{fig:abla-train}  (b) and (c)} details the impact of our deformable augmentation module during training. We employ the optimal data ratio from Panel (a) to evaluate the effects. Adding deformation to only the source domain (``Source") or the target domain (``Target") leads to a decrease in DSC score but an increase in NSD, indicating enhanced handling of diverse boundary conditions. However, applying deformations to both domains (``Source+Target") results in performance degradation, likely due to the complexity added by excessive augmentation, which could hinder model learning. This panel confirms that our deformation augmentation module contributes to scaling up training data with varied templates, albeit with nuances in application.

\textbf{Figure~\ref{fig:abla-train}  (d) and (e)} presents the outcomes of utilizing different qualities of pseudo labels in training. Settings from the best-performed experiments in Panel (a), including a 1:1 data ratio and source domain label deformations, were used. The ``low quality", ``middle quality'', and ``high quality" category indicates the pseudo labeling results on the target domain training data with PL~\cite{yan2022unsupervised}, ASC~\cite{xu2023asc}, and ours ASCPlus, respectively. Results demonstrate that superior pseudo-labels can lead to better overall performance on diffuse UDA.

\textbf{Figure~\ref{fig:abla-train}  (f)} pays attention to the training process of our ASCPlus model, we can observe that our method is not benefit from the deformable data augmentation during the training process.

\begin{figure}[!tbp]
\includegraphics[width=1\linewidth]{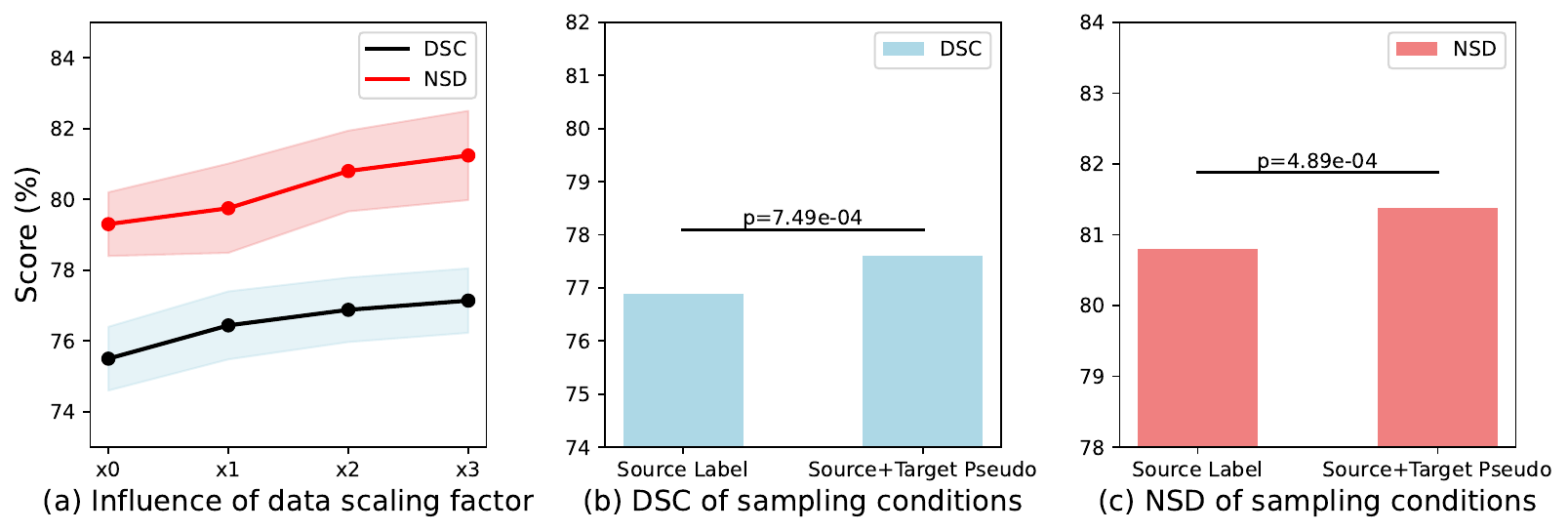}
\centering
\caption{Detailed analysis on using different types of testing data in our diffuse UDA framework. (a) a different number of samples to train the neural network. (b) different sampling strategy to get the training data.}
\label{fig:abla-infer}
\end{figure}

\begin{figure}[!tbp]
\includegraphics[width=1\linewidth]{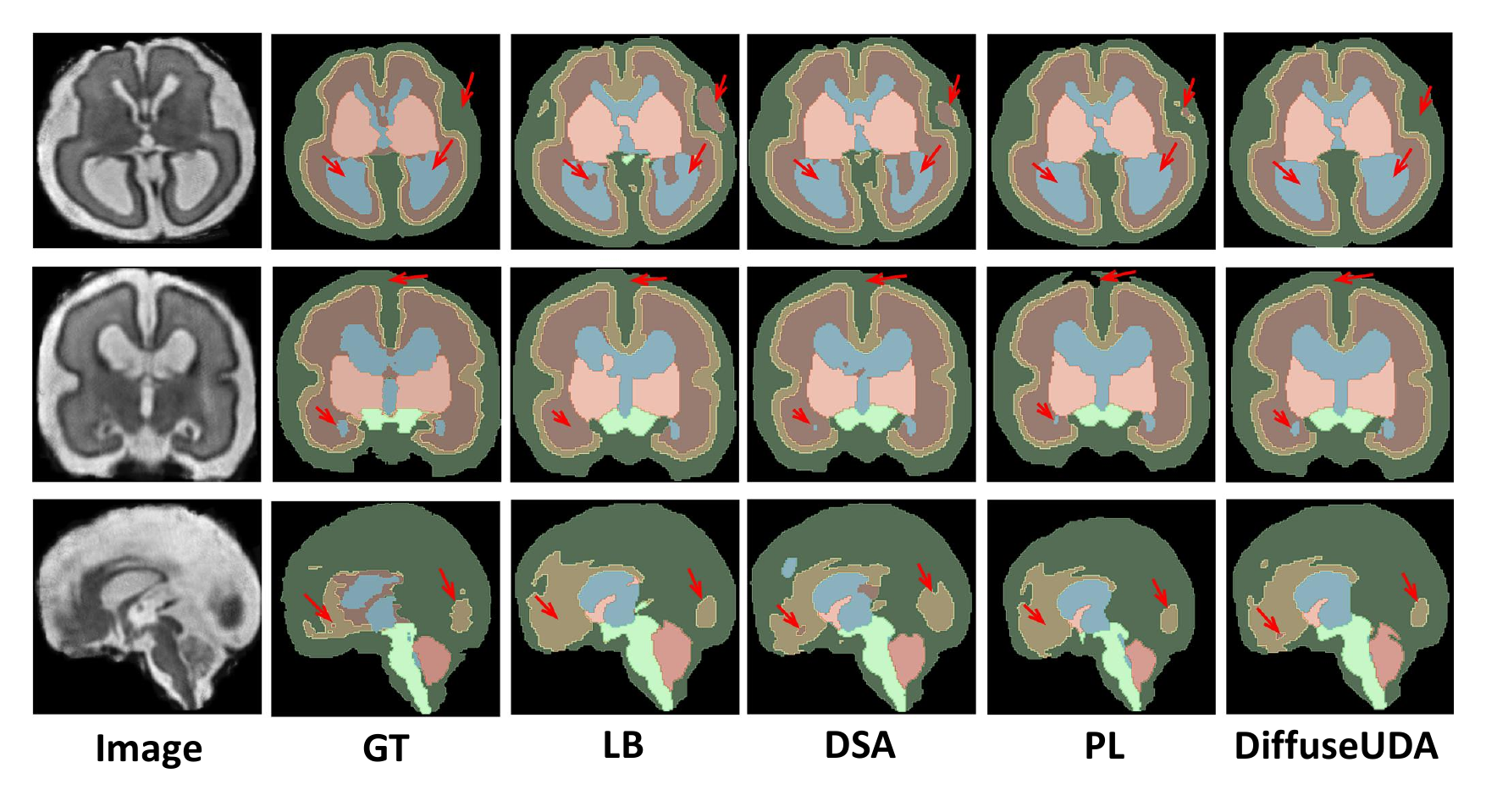}
\centering
\caption{The qualitative comparison of our method and other methods on the tissue segmentation dataset with other advanced methods. The highlighted regions are marked by the red arrow.}
\label{fig:qualitative}
\end{figure}

\noindent \textbf{Analysis on generation process for Diffuse-UDA}
Figure~\ref{fig:abla-infer} presents results from two key investigations into our model's training and performance. Panel (a) of Figure~\ref{fig:abla-infer} illustrates the impact of scaling up training data. Here, `x0' indicates training using only the source domain data. `x1' represents a scenario where both source and target domain data are used, with the number of generated samples equal to that of the source domain data. `xn' denotes the addition of $n$ times the source data quantity to the training set, exploring the effects of increasing training data volume. Panel (b) of Figure~\ref{fig:abla-infer} explores the influence of using different types of labels for image sampling. The results indicate that applying a deformation operation to both source and target domain data significantly enhances model performance, showcasing the benefits of this approach in handling diverse data characteristics.

\subsection{Conclusion}
In conclusion, Diffuse-UDA represents a significant advancement in the field of medical image segmentation by addressing the domain gap between datasets from varying resource environments. By leveraging diffusion models to generate high-quality image-mask pairs tailored to target domain characteristics, Diffuse-UDA enhances the performance of UDA tasks. Our comprehensive evaluations demonstrate that Diffuse-UDA not only outperforms existing UDA and semi-supervised methods but also approaches or exceeds the performance of models trained directly on target domain data. This novel approach has the potential to revolutionize the deployment of AI systems in medical imaging, offering a more fair and effective diagnostic tool that can improve outcomes, save time, and reduce human error across diverse healthcare settings.

\section{Methods}

Within the UDA paradigm, we define the source domain as $D_{s} = \{ (x_{s}^i, y_{s}^i) \}_{i=1}^{M}$, where $x_{s}^i$ represents the source images and $y_{s}^i$ their corresponding annotations. The target domain is denoted by $D_{t} = \{ x_{t}^j \}_{j=1}^{N}$, which consists of images from the FeTA benchmark devoid of annotations. The primary objective is to harness the labeled data from $D_{s}$ alongside the unlabeled data from $D_{t}$ to train a semantic segmentation model capable of accurate performance on the target domain.
To bridge the domain gap, we introduce the Diffuse-UDA framework, which synthesizes a novel dataset $D_{g} = \{ (x_{g}^k, y_{g}^k) \}_{k=1}^{K}$, amalgamating features from both $D_{s}$ and $D_{t}$. The model is subsequently trained using the labeled data from $D_{s}$ and the pseudo-labeled data from $D_{g}$, enhancing the model's adaptability to $D_{t}$. Figure~\ref{fig:pipeline} provides a graphical overview of our proposed methodology.

\subsection{ASCPlus: Enhanced appearance and structure consistency model for UDA in medical image segmentation}
We propose the UDA framework, ASCPlus, designed to align the appearance and structure gaps between different domains. To adapt the segmentation model to the varying appearances across domains, we enforce consistency before and after a frequency-based image transformation that swaps appearances between source and target domain data. Recognizing that even within the same domain, anatomical structures can exhibit significant variability across individuals, we further enhance the model's adaptability to structural variations in the target domain by encouraging prediction consistency under different structural perturbations.
 
\subsubsection{Appearance consistency learning via adaptive frequency-based transformation}

Domain shifts between the source and target domain data primarily arise from variations in texture, different hospital sensors, illumination, and other low-level sources of variability. Traditional UDA methods employing GANs~\cite{al2021olva} for synthetic style-transfer images often fail to capture such domain shifts effectively. To address this, we align the low-level statistics based on Fourier transformation to narrow the distribution gap between the two domains. Specifically, for source data, we compute the Fast Fourier Transform (FFT) of each input image to obtain an amplitude spectrum $\mathcal{F}^{A}$ and a phase component $\mathcal{F}^{P}$. The low-frequency part of the amplitude of the source image $\mathcal{F}^{A}(x_{s})$ is swapped with the amplitude of the target image $\mathcal{F}^{A}(x_{t})$. The transformed spectral representation of $x_{s}$, along with the original phase $\mathcal{F}^{P}(x_{s})$, is then mapped back to the image $x_{sft}$ using the inverse FFT (iFFT). As a result, $x_{sft}$ retains the content of $x_{s}$ while adopting the appearance of $x_{t}$. This process is formally defined as:

\begin{equation}
\mathcal{F}^{A}(x_{sft})=\underline{M} \cdot \mathcal{F}^{A}(x_{t}) + (1-\underline{M}) \cdot \mathcal{F}^{A}(x_{s}),
\label{eq:1}
\end{equation}

\begin{equation}
x_{sft} = \mathcal{F}^{-1}([\mathcal{F}^{A}(x_{sft}), \mathcal{F}^{P}(x_{s})]),
\label{eq:2}
\end{equation}
where the mask $\underline{M} = \mathcal{I}_{(h,w,d) \in [-\beta H : \beta H, -\beta W : \beta W, -\beta D : \beta D]}$ controls the proportion of the swapped part over the entire amplitude based on the parameter $\beta \in (0,1)$. Here, we assume the center of the image is at (0, 0, 0). To adaptively select the value of $\beta$, we calculate the grayscale histograms of the source and target domain images and compute the Euclidean distance between them. This distance determines $\beta$: a larger distance requires more information to be swapped, resulting in a larger $\beta$. Considering the variability of $\beta$ within the domain, we set a minimum value of $\beta$ to 0.1 and multiply the final $\beta$ value by a factor between 0.5 and 1.5. Specifically, $\beta$ is determined as follows:

\begin{equation}
\beta = \max(0.1, \frac{d_{hist}(x_{s}, x_{t})}{d_{max}} \cdot \alpha),
\label{eq:beta}
\end{equation}
where $d_{hist}(x_{s}, x_{t})$ is the Euclidean distance between the histograms of the source and target images, $d_{max}$ is the maximum possible distance, and $\alpha$ is a factor between 0.5 and 1.5. Let $P_{s}$ and $P_{sft}$ be the predictions of $x_{s}$ and $x_{sft}$, respectively. We train a student network with domain-aligned images $x_{sft}$, the original images $x_{s}$, and the labels $y_{s}$ by minimizing the dice loss:

\begin{equation}
L_{seg} = L_{dice}(P_{s}, y_{s}) + L_{dice}(P_{sft}, y_{s}).
\label{eq:3}
\end{equation}

This loss function imposes implicit regularization before and after the frequency-based transformation, ensuring that the source domain image $x_{s}$ and its transformed image $x_{sft}$ yield consistent predictions. In UDA settings, the target domain images lack labels. To address this, we propose a teacher model to maintain semantic consistency across domain transformations. Specifically, the target domain image $x_{t}$ and its aligned image $x_{sft}$ are viewed as representations of an object under different domains. Given inputs $x_{sft}$ for both teacher and student models, we expect consistent predictions. To mitigate the potential semantic information loss caused by appearance transformation, we employ dual consistency, directing the model to focus on invariant information between the two views. Denoting the outputs of the student model as $f(\cdot)$ and the teacher model as $f'(\cdot)$, we calculate the appearance consistency loss $L^{app}_{con}$ as:

\begin{equation}
L^{app}_{con} = \frac{1}{N} \sum^{N}_{i=1} ||f(x_{t,i}) - f'(x_{tfs,i})||^2 + \frac{1}{N} \sum^{N}_{i=1} ||f(x_{tfs,i}) - f'(x_{t,i})||^2.
\label{eq:4}
\end{equation}

\subsubsection{Structure consistency learning via mixup}

Although frequency-based transformation and appearance consistency align the styles of the two domains, the variance in tissue structure in pathological subjects still poses a challenge to domain alignment, limiting the model's generalization ability. To address this, we utilize a teacher-student model to maintain prediction consistency $L^{str}_{con}$ under structure perturbation $sp$. Here, $sp$ refers to the CutMix~\cite{yun2019cutmix} technique, a type of mixup technology~\cite{zhang2018mixup,gong2022vqamix} for data augmentation. CutMix augments training images by cutting and pasting patches among training samples, which helps the model generalize better by exposing it to a variety of structure perturbations.

Since appearance transformation does not affect structure information, we apply $sp$ to both $x_{t}$ and $x_{tfs}$ to obtain $x_{t,sp}$ and $x_{tfs,sp}$, respectively. The CutMix operation involves randomly selecting a rectangular region in the image and replacing it with a patch from another image. These perturbed images are then fed into the teacher-student model, and their predictions are expected to be consistent. The structure consistency loss $L^{str}_{con}$ and the appearance consistency loss $L^{app}_{con}$ are combined into a single loss term $L_{asc}$ to ensure consistency across both appearance and structural perturbations:

\begin{equation}
L_{asc} = \frac{1}{N}\sum^{N}_{i=1}||f(x_{t,i,sp}) - f'(x_{tfs,i})||^2 + \frac{1}{N}\sum^{N}_{i=1}||f(x_{tfs,i,sp}) - f'(x_{t,i})||^2.
\label{eq:5}
\end{equation}

We calculate appearance and structure consistency using the same teacher model. Its model weights $\theta'$ are updated using the exponential moving average (EMA) of the student model $f(\theta)$, i.e.,

\begin{equation}
\theta'_{t} = \gamma\theta'_{t-1} + (1 - \gamma)\theta_{t},
\end{equation}
where $\gamma$ is the EMA decay rate that reflects the influence level of the current student model parameters. The use of EMA helps in stabilizing the training process and improves the robustness of the teacher model. Let $\lambda$ control the trade-off between the supervised loss $L_{seg}$ and the consistency loss $L_{asc}$, the overall loss is defined as:

\begin{equation}
L_{all} = L_{seg} + \lambda L_{asc}.
\end{equation}

\subsection{Pseudo labeling with ASCPlus}

To synthesize images that are stylistically aligned with the target domain, under the guidance of corresponding masks, we first generate pseudo labels for the images $x_t$ from the target domain $D_t$. This is achieved by deploying a segmentation model initially trained on the source domain dataset $D_s$, or a UDA model $f_{\theta}$ trained on $D_s$ and $D_t$. Pseudo labeling involves using the model's predictions as labels for the unlabeled target domain data, which helps in leveraging the target domain data for further training.

To obtain high-quality pseudo labels, we use the UDA model $f_{\theta}$ to predict the pseudo label $y_t$ for each image $x_t \in D_t$, which is formalized as:
\begin{equation}
y_t = f_{\theta}(x_t).
\end{equation}

The quality of the pseudo labels is crucial for the performance of the model. Hence, we adopt confidence thresholding, where only the predictions with high confidence are considered as pseudo labels. Using these high-confidence pseudo labels produced by $f_{\theta}$, we can subsequently condition the generation of the dataset $D_g$, utilizing the image-mask pairs from both $D_s$ and the pseudo-labeled $D_t$. This synthetic dataset $D_g$ aids in better aligning the source and target domains, improving the model's generalization capability. Moreover, we also ensemble the results of last 10 epoch to get better pseudo label. By integrating these strategies, ASCPlus effectively narrows the domain gap by addressing both appearance and structure inconsistencies, leveraging pseudo labeling to enhance the alignment further. This comprehensive approach ensures robust performance across varied and challenging target domains.

\subsection{Conditional diffusion model are robust generator for UDA}

In the realm of UDA, leveraging robust feature representations that can generalize across different domains is crucial. To this end, we propose a novel approach utilizing a conditional diffusion model to act as a robust generator for domain-adapted image segmentation. Our design is simple: We fuse image-label pairs from the source domain with image-pseudo-label pairs from the target domain to train a diffusion model that generates images with segmentation coherent with the target domain style. Our method is motivated by the following principles:

\textbf{Implicit denoising learning.} At its core, the diffusion model is a generative model that crafts data by iteratively removing noise. When dealing with pseudo-labeled data imbued with noise, the model must discern how to reconstruct clean images from such noisy conditional inputs (i.e., unreliable pseudo-label). This learning process echoes the principles of denoising autoencoders, compelling the model to capture robust feature representations post-denoising. These representations are potentially more conducive to the performance of downstream tasks. By conditioning the diffusion process on noisy pseudo labels, we implicitly guide the model to learn these noise-invariant features. Beyond that, the original clear image-label pair can also guide for reconstructing of the image from noisy labels.

\textbf{Feature space smoothing.} Introducing noise through pseudo labels during training can enhance the model's robustness to noise. The diffusion model is challenged not only to learn the correct data distribution but also to disregard inconsistencies from pseudo-label noise. Such training primes the model for greater resilience against real-world noise in the target domain, such as scanning artifacts and blurring, and the noisy condition (i.e., pseudo label) can act like the regularizer to avoid the model overfitting to the source domain data. The generated image-pseudo label pair can work similarly to the mixup operation to avoid the model overfitting in the downstream training task.

By integrating image-label pairs from the source domain with image-pseudo-label pairs from the target domain, our conditional diffusion model is jointly trained to generate images with target domain style segmentation. This approach not only serves to align the data distributions between the domains but also imbues the model with the robustness necessary for effective UDA. The conditional diffusion model thus emerges as a potent tool for domain adaptation, adept at navigating the challenges posed by domain discrepancies and noise.

\subsubsection{Structure of conditional diffusion model}
In this work, we explore the Denoising Diffusion Probabilistic Model (DDPM)~\cite{ho2020denoising} for domain adaptation tasks, specifically addressing the unique challenges posed by fetal brain imaging. Given the significant differences between the structures of fetal brains and typical medical or natural images, we initially perform zero-cropping and resample the fetal images to achieve uniformity in size.

In the standard DDPM model, the forward diffusion process denoted as $q$, incrementally introduces Gaussian noise $\epsilon \sim \mathcal{N}(0, \mathbf{I})$ to an image $x_0$ from the training dataset. This process is governed by a variance schedule $\bar{\alpha}_t$, across a predefined number of timesteps $\underline{T}$. We employ a cosine noise schedule to ensure stable noise generation. The noisy sample at timestep $\underline{t}$ is represented as follows:
\begin{equation}
x_{\underline{t}} = \sqrt{\bar{\alpha}_{\underline{t}}} x_0 + \sqrt{1 - \bar{\alpha}_{\underline{t}}} \epsilon, \quad \text{for } 1 < {\underline{t}} \leq {\underline{T}}.
\end{equation}

To synthesize meaningful images, such as pathological MRIs of fetal brains where different brain parts occupy varying spatial ratios, we utilize a mask to guide the generation process. The conditioning of the image generation involves encoding the condition $c$ (either a mask or a predicted pseudo mask) using a linear layer $f_e$. The encoded condition is then concatenated with the input image $x_{\underline{t}}$ to form the final conditioning input, as shown below:
\begin{equation}
\hat{x}_{\underline{t}} = [x_{\underline{t}}, f_e(c)],
\label{embedding}
\end{equation}
where $[.,.]$ denotes point-wise addition.

The conditioning training process, illustrated in Fig. 3, involves concatenating the segmentation mask $c$ with the noisy image $x_{\underline{t}}$ at each timestep $t$. The denoising process, also known as generative sampling, is defined as:
\begin{equation}
x_{\underline{t-1}} = \frac{1}{\sqrt{\alpha_{\underline{t}}}} \left(x_{\underline{t}} - \frac{1 - \alpha_{\underline{t}}}{\sqrt{1 - \bar{\alpha}_{\underline{t}}}} \epsilon_{\theta}(\hat{x}_{\underline{t}}, {\underline{t}})\right) + \sigma_{\underline{t}} z,
\label{eq:denosing}
\end{equation}
where $z \sim \mathcal{N}(0, \mathbf{I})$, $\sigma_{\underline{t}} = \sqrt{\beta_{\underline{t}}}$, $\beta_{\underline{t}} \in (0, 1)$, and $\epsilon_{\theta}$ is the noise predicted by a U-Net model. To ensure the reliability of labels used for generation, we exclusively utilize masks from the source domain $D_t$ to generate image-mask pairs $D_g$. Let $i$ be the index of voxels before and after reconstruction, we use the L1 loss following~\cite{kim2022diffusion} to effectively train the diffusion model:
\begin{equation}
L_{\text{MAE}} = \frac{1}{n} \sum_{i=1}^{n} |\epsilon_i - \epsilon'_i|.
\end{equation}

\subsubsection{Deformable augmentation enables scaling-up and diverse generation}
Given the considerable variability in the structures of fetal brains and tumors in real-world scenarios, using identical sample images from the same mask may curtail the diversity of synthetic samples. Additionally, in cases of fetal brain disorders, the affected brain tissue might exhibit variations in spatial extent and positioning. To address these challenges, we have developed a PyTorch-based deformation module, $f_{\text{deform}}$, which facilitates rapid deformable augmentation using CUDA. This module integrates both affine and elastic transformations, and its implementation is detailed in Algorithm 1.

To ensure stability in the training process, given the inherent noise in the pseudo-labels, deformation augmentation is applied solely to the image-mask pairs from the dataset $D_s$. During the sampling phase, the deformation augmentation is extended to both the labels from $D_s$ and the pseudo-labels from $D_t$, enhancing the robustness and diversity of the generated samples. Consequently, the training process as described in Equation~\ref{eq:denosing} is modified as follows:
\begin{equation}
    x_{\underline{t-1}} = \frac{1}{\sqrt{\alpha_{\underline{t}}}} \left(x_{\underline{t}} - \frac{1 - \alpha_{\underline{t}}}{\sqrt{1 - \bar{\alpha}_{\underline{t}}}} \epsilon_{\theta}(\hat{x}_{\underline{t}}, \underline{t})\right) + \sigma_{\underline{t}} z,
    \label{eq:denosing}
\end{equation}
where the generation of the conditional image is reformulated in Equation~\ref{embedding}:
\begin{equation}
    \hat{x}_t = [x_t, f_e(f_{\text{deform}}(c))],
\end{equation}
and the process for reconstructing the training data is updated as:
\begin{equation}
    x_{\underline{t}} = \sqrt{\bar{\alpha}_{\underline{t}}} f_{\text{deform}}(x_0) + \sqrt{1 - \bar{\alpha}_{\underline{t}}} \epsilon.
\end{equation}
These modifications leverage the flexibility of deformable augmentation to simulate a wider range of anatomical variations, thereby enhancing the model's ability to generalize across different and challenging datasets.

\subsection{Implementation}
Our models were developed using the PyTorch 2.0.1 framework and trained on NVIDIA A800 GPUs equipped with CUDA 12.1, each providing 80GB of memory. Following the FeTA2021 competition's winning approach~\cite{payette2021automatic,payette2023fetal}, we employed SegResNet\cite{myronenko20193d} as the backbone network. For training the diffusion model, we set the batch size to 1 and used a UNet with GroupNorm as the backbone, training it for 100 epochs with the Adam optimizer and setting the sampling steps to 250. For the downstream task, we utilized the Adam optimizer with an initial learning rate of $1 \times 10^{-4}$ and a batch size of 4, equally split between domains, over 100 epochs. To ensure robustness, we reported the mean outcomes from three separate trials for all experiments. Following~\cite{xu2023asc}, we pre-processed and resized images and masks to a uniform dimension of $128 \times 128 \times 128$ voxels for fetal brain segmentation task. All the results are obtained by three individual runs.
The specific training parameters for the Diffusion model are as follows: a batch size of 1, 100,000 iterations, an input size of 128x128, 64 channels, 1 residual block, 250 timesteps, and samples are saved every 1000 iterations. Additionally, the training involves gradient accumulation every 2 steps, an exponential moving average (EMA) decay setting of 0.995, and the optimizer used is Adam with a learning rate 1e-5. In the sampling phase of the Diffusion model, real labels from the source domain and pseudo-labels from the target domain are used as conditions for the sampling process.

\subsection{Evaluation metrics}
For the evaluation of our segmentation algorithms, we employ three metrics as suggested by recent studies~\cite{xu2023asc,Reinke2023UnderstandingMP,wang2024towards}: the dice similarity coefficient (DSC), the normalized surface dice (NSD), and average surface distance (ASD). The DSC measures the overlap between the predicted and ground truth segmentation masks, providing a direct quantification of the segmentation accuracy. The NSD and ASD, on the other hand, evaluates the conformity of the surface contours of the predicted segmentation to those of the actual tumor, which is particularly crucial for accurately delineating the irregular and complex boundaries of metastatic brain tumors.

\clearpage
\begin{appendices}

\begin{algorithm}
\caption{Pseudo Code of Deformable Transformation}\label{alg:affine_elastic_3d_gpu}
\begin{algorithmic}[1]
\Procedure{DeformableTransform}{$x$: voxel, $r$: rotation matrix, $sc$: scale, $sh$: shift, $w$ : window size, $f$ : field size, $\alpha$: displacement fields scaling factor, $sn$ : smooth number}
    \State $A: Affine Matrix$
    \State $A \gets \Call{AngleAxisToRotationMatrix}{r}$
    \For{each dimension $d$ in $\{0, 1, 2\}$}
        \State $A[:, d, 3] \gets sh[:, d]$
        \For{each axis $a$ in $\{0, 1, 2\}$}
            \State $A[:, d, a] \gets A[:, d, a] \times sc[:, d]$
        \EndFor
    \EndFor
    \State $A \gets A[:, 0:3, :]$ 
    \State $grid \gets \Call{AffineGrid}{A, x[0].size()}$
    \State $pad \gets \Call{CalculatePadding}{w}$
    \State Initialize displacement fields $dz$, $dy$, $dx$ with $w$ and $f$
    \State Scaling $dz$, $dy$, $dx$ using $\alpha$
    \For{$i \gets 1$ \textbf{to} $sn$}
        \State $dz \gets \Call{Smooth with Conv}{dz, w}$
        \State $dy \gets \Call{Smooth with conv}{dy, w}$
        \State $dx \gets \Call{Smooth with conv}{dx, w}$
    \EndFor
    \State Crop padding from displacement fields $dz$, $dy$, $dx$
    \State Resize $dz$, $dy$, $dx$ to match the size of input volumes
    \State Add $dz$, $dy$, $dx$ to the $grid$ to create the final transformation grid
    \State $x_{affine} \gets \Call{GridSample}{x, grid}$
    \State \textbf{return} $x_{affine}$
\EndProcedure
\end{algorithmic}
\end{algorithm}
\newpage
\clearpage
\clearpage
\clearpage

\clearpage

\begin{table}[]
\centering
\caption{Comparison with the state-of-the-art methods on the FeTA benchmark. The first column denotes the group of methods including registration-based method  (REG), UDA and semi-supervised learning (SSL). ``UB'' and ``LB'' denote the upper bound and lower bound, respectively. The best results of each category of method are shown in \textbf{bold}.}
\begin{tabular}{@{}cccccc}
\toprule
\multirow{2}{*}{Type} & \multirow{2}{*}{Method}  & \multicolumn{2}{c}{Abnormal Fetal Brain}& \multicolumn{2}{c}{Normal Fetal Brain}\\ 
     &        & DSC         & NSD         & DSC         & NSD         \\ \midrule
UB   & $D_{t}$ supervise & $81.1_{\pm0.2}$& $83.0_{\pm0.2}$& $83.6_{\pm0.3}$& $87.8_{\pm0.4}$\\
LB   & $D_{s}$ supervise & $73.1_{\pm0.6}$ & -& $79.0_{\pm0.3}$& -\\ \midrule
 REG & SCALE~\cite{sanroma2018learning}& $63.6_{\pm1.5}$& -& $77.3_{\pm1.7}$&-\\ \midrule
     & FDA~\cite{yang2020fda} & $74.4_{\pm0.4}$& -& $80.4_{\pm0.4}$& -\\
UDA  & OLVA~\cite{al2021olva} & $73.3_{\pm0.6}$& -& $79.1_{\pm0.3}$& -\\
 & DSA~\cite{han2021deep} & $73.4_{\pm0.6}$& -& $79.9_{\pm0.4}$&-\\
     & ASC~\cite{xu2023asc}& $76.6_{\pm0.1}$& $76.4_{\pm0.3}$& $81.5_{\pm0.1}$& $84.3_{\pm0.2}$\\ \midrule
     & Cutmix~\cite{yun2019cutmix} & $74.1_{\pm0.1}$& -& $79.7_{\pm0.5}$& -\\
SSL  & ASE~\cite{lei2022semi} & $73.7_{\pm0.3}$& -& $79.8_{\pm0.3}$& -\\
     & PL~\cite{yan2022unsupervised} & $74.3_{\pm0.4}$ & $75.1_{\pm0.6}$& $79.1_{\pm0.2}$& $79.5_{\pm0.3}$\\ 
     & DiffSyn~\cite{dorjsembe2023conditional} & $73.8_{\pm0.7}$ & $73.4_{\pm0.2}$& $79.3_{\pm0.6}$ & $79.9_{\pm0.4}$\\
     \midrule
     Ours& DiffuseUDA & $\bm{81.3_{\pm0.4}}$& $\bm{84.0_{\pm0.7}}$& $\bm{84.5_{\pm0.5}}$& $\bm{88.8_{\pm0.9}}$\\ \midrule
      \multirow{2}{*}{\textit{p-value}}& Ours v.s. UB & 0.482& 0.076& 0.056& 0.153\\
      & Ours v.s. 2nd Method& \textless0.005& \textless0.005& \textless0.005& \textless0.005\\
     \bottomrule
\end{tabular}
\label{exp:sota}
\end{table}

\begin{table}[]
\caption{Comparison with the state-of-the-art methods on the MMWHS benchmark. The first column denotes the group of methods including UDA and semi-supervised learning (SSL). ``UB'' and ``LB'' denote the upper bound and lower bound, respectively. The best results of each category of method are shown in \textbf{bold}.}
\begin{tabular}{cccccc}
\hline
\multirow{2}{*}{Type}        & \multirow{2}{*}{Method}      & \multicolumn{2}{c}{CT to MR}& \multicolumn{2}{c}{MR to CT}\\
\multicolumn{1}{c}{}            &            & DSC     & ASD    & DSC     & ASD    \\ \hline
\multicolumn{1}{c}{UB} & $D_{t}$ supervise           & $84.5_{\pm1.6}$   & $2.1_{\pm0.4}$   & $87.5_{\pm0.5}$& $1.5_{\pm0.6}$   \\
\multicolumn{1}{c}{LB} & $D_{s}$ supervise         & 21.1    & 24.8 & 38.5   & 15.2 \\ \hline
\multirow{10}{*}{UDA}            & PNP-AdaNet~\cite{dou2019pnp} & 63.9    & 8.9    & 63.9    & 12.8   \\
                                & AdaOutput~\cite{tsai2018learning}  & 51.9    & 5.7    & 59.9    & 9.6    \\
                                & CycleGAN~\cite{zhu2017unpaired}   & 50.7    & 6.6    & 57.6    & 10.8   \\
                                & CyCADA~\cite{hoffman2018cycada}     & 57.5    & 7.9    & 64.4    & 9.4    \\
                                & SIFA~\cite{chen2020unsupervised}       & 63.4    & 5.7    & 74.1    & 7.0    \\
                                & DSFN~\cite{zou2020unsupervised}       & -       & -      & 75.8    & -      \\
                                & DSAN~\cite{han2021deep}       & 66.5    & 5.4    & 78.5    & 5.9    \\
                                & LMISA-3D~\cite{jafari2022lmisa}   & 70.8    & 3.6    & 81.5    & 2.3    \\
                                & FSUDA~\cite{liu2023reducing}   & 70.6    & 4.1    & 85.0    & 2.6    \\              
                                & FSUDAv2~\cite{liu2023structure}   & 75.5    & 3.3    & 86.1    & 2.4    \\

                                & MAPSeg~\cite{zhang2024mapseg}   & 80.3    & -    & -    & -    \\
                                \hline
\multirow{3}{*}{SSL}                                
& SS-Net~\cite{wu2022exploring}     & 60.2    & 5.9    & -       & -      \\
                                & BCP~\cite{bai2023bidirectional}        & 64.1    & 4.5    & -       & -      \\ 
&  A\&D~\cite{wang2024towards}       & $71.4$    & 7.9    & $86.6_{\pm0.9}$    & $3.5_{\pm1.0}$    \\ \midrule
Ours&  DiffuseUDA       & $\bm{84.1_{\pm0.6}}$    & $\bm{2.0_{\pm0.2}}$    & $\bm{88.3_{\pm0.3}}$    & $\bm{1.6_{\pm0.1}}$    \\ \midrule
\multirow{2}{*}{\textit{p-value}}& Ours v.s. UB & 0.665& 0.675& 0.076& 0.789\\
& Ours v.s. 2nd Method& \textless0.005& \textless0.005& 0.036& \textless0.005\\
\bottomrule
\end{tabular}
\end{table}

\begin{table}[tbp]
\centering
\caption{Ablation study of the components in our proposed Diffuse-UDA generation on the feta dataset with the pseudo label setting. ``Source mask'' and ``Target mask'' indicates use the source domain segmentation and target domain pseudo label segmentation for sampling the image-mask pair. ``Deformable'' indicates apply the deformable augmentation to the masks. ``Scale-up'' indicates sample the mask with additional 1 times. M1 to M5 only use half of the samples in the source domain and target domain training set, while M6 use all the samples in the source domain training set and target domain training set, and sample the mask for 4 times.}
\begin{tabular}{@{}ccccccc@{}}
\toprule 
Method&  Source mask &  Target mask &   Deformation&  Scale-up & Full training set &
DSC\\ 
\hline
M1&&       &     & & &$75.3_{\pm0.5}$\\ 
M2&$\surd$&&       &  & &$75.8_{\pm0.4}$\\ 
M3&$\surd$&$\surd$&& & &$76.6_{\pm0.3}$\\ 
M4&$\surd$&$\surd$&$\surd$& & &$77.0_{\pm0.2}$\\
M5&$\surd$&$\surd$&$\surd$&$\surd$&  &$78.2_{\pm0.2}$\\ \hline
M6&$\surd$&$\surd$&$\surd$&$\surd$& $\surd$ &$81.1_{\pm0.2}$\\ \hline

\end{tabular}
\label{tab:abla-uda}
\end{table}
\clearpage

\begin{table}[]
\centering
\caption{Ablation study of the contribution of each component in the proposed ASCPlus. Each result are obtain by averaging the results of three different runs. ``AdaFFT'' indicates the adaptive setting of $\beta$ in Eq.3. ``Ensemble'' indicates ensemble the segmentation results of last 10 epochs.}
\begin{tabular}{@{}ccccccccc@{}}
\toprule
\multirow{2}{*}{Method} & \multirow{2}{*}{$L_{seg(x_{s})}$} & \multirow{2}{*}{$L_{seg(x_{sft})}$} & \multicolumn{3}{c}{$L_{asc}$}                                       & \multirow{2}{*}{AdaFFT} & \multirow{2}{*}{Ensemble} & \multirow{2}{*}{Avg DSC} \\ \cmidrule(lr){4-6}
                        &                                   &                                     & $L^{app}_{con(x_{t})}$ & $L^{app}_{con(x_{tfs})}$ & $L^{str}_{con}$ &                               &                           &                          \\ \midrule
M1                      & $\surd$                           &                                     &                        &                          &                 &                               &                           & $75.3_{\pm0.5}$          \\
M2                      & $\surd$                           & $\surd$                             &                        &                          &                 &                               &                           & $76.6_{\pm0.2}$          \\
M3                      & $\surd$                           & $\surd$                             & $\surd$                &                          &                 &                               &                           & $77.3_{\pm0.3}$          \\
M4                      & $\surd$                           & $\surd$                             & $\surd$                & $\surd$                  &                 &                               &                           & $77.9_{\pm0.2}$          \\
M5                      & $\surd$                           & $\surd$                             & $\surd$                & $\surd$                  & $\surd$         &                               &                           & $78.5_{\pm0.1}$          \\ 
M6                      & $\surd$                           & $\surd$                             & $\surd$                & $\surd$                  & $\surd$         & $\surd$                       &                           & $78.6_{\pm0.2}$          \\ 
M7                      & $\surd$                           & $\surd$                             & $\surd$                & $\surd$                  & $\surd$         & $\surd$                       & $\surd$                   & $78.9_{\pm0.1}$          \\ \bottomrule
\end{tabular}
\label{tab:abla-asc}
\end{table}

\end{appendices}


\begin{thebibliography}{200}
\bibitem{al2021olva} Al Chanti, Dawood and Mateus, Diana, ``OLVA: Optimal Latent Vector Alignment for Unsupervised Domain Adaptation in Medical Image Segmentation", in \textit{MICCAI}, Springer, pp. 261--271, 2021.
\bibitem{bai2023bidirectional} Bai, Yunhao and Chen, Duowen and Li, Qingli and Shen, Wei and Wang, Yan, ``Bidirectional copy-paste for semi-supervised medical image segmentation", in \textit{Proceedings of the IEEE/CVF conference on computer vision and pattern recognition}, , pp. 11514--11524, 2023.
\bibitem{basak2023pseudo} Basak, Hritam and Yin, Zhaozheng, ``Pseudo-label guided contrastive learning for semi-supervised medical image segmentation", in \textit{Proceedings of the IEEE/CVF conference on computer vision and pattern recognition}, , pp. 19786--19797, 2023.
\bibitem{chen2020unsupervised} Chen, Cheng and Dou, Qi and Chen, Hao and Qin, Jing and Heng, Pheng Ann, ``Unsupervised bidirectional cross-modality adaptation via deeply synergistic image and feature alignment for medical image segmentation", \textit{TMI}, vol. 39, no. 7, pp. 2494--2505, 2020.
\bibitem{chung2022score} Chung, Hyungjin and Ye, Jong Chul, ``Score-based diffusion models for accelerated MRI", \textit{Medical image analysis}, vol. 80, no. , pp. 102479, 2022.
\bibitem{de2022adverse} De Asis-Cruz, Josepheen and Andescavage, Nickie and Limperopoulos, Catherine, ``Adverse prenatal exposures and fetal brain development: insights from advanced fetal magnetic resonance imaging", \textit{Biological Psychiatry: Cognitive Neuroscience and Neuroimaging}, vol. 7, no. 5, pp. 480--490, 2022.
\bibitem{dorjsembe2022three} Dorjsembe, Zolnamar and Odonchimed, Sodtavilan and Xiao, Furen, ``Three-dimensional medical image synthesis with denoising diffusion probabilistic models", in \textit{Medical Imaging with Deep Learning}, , pp. , 2022.
\bibitem{dorjsembe2023conditional} Dorjsembe, Zolnamar and Pao, Hsing-Kuo and Odonchimed, Sodtavilan and Xiao, Furen, ``Conditional Diffusion Models for Semantic 3D Medical Image Synthesis", \textit{arXiv preprint arXiv:2305.18453}, vol. , no. , pp. , 2023.
\bibitem{dou2019pnp} Qi Dou, Cheng Ouyang, Cheng Chen, Hao Chen, Ben Glocker, Xiahai Zhuang, and Pheng-Ann Heng, ``Pnp-adanet: Plug-and-play adversarial domain adaptation network at unpaired cross-modality cardiac segmentation,'' \textit{IEEE Access}, vol. 7, pp. 99065--99076, 2019.
\bibitem{fan2023survey} Fan, Yuheng and Liao, Hanxi and Huang, Shiqi and Luo, Yimin and Fu, Huazhu and Qi, Haikun, ``A Survey of Emerging Applications of Diffusion Probabilistic Models in MRI", \textit{arXiv preprint arXiv:2311.11383}, vol. , no. , pp. , 2023.
\bibitem{fidon2022spatio} Fidon, Lucas and Viola, Elizabeth and Mufti, Nada and David, Anna L and Melbourne, Andrew and Demaerel, Philippe and Ourselin, S{\'e}bastien and Vercauteren, Tom and Deprest, Jan and Aertsen, Michael, ``A spatio-temporal atlas of the developing fetal brain with spina bifida aperta", \textit{Open Research Europe}, vol. 1, no. , pp. 123, 2022.
\bibitem{gholipour2017normative} Gholipour, Ali and Rollins, Caitlin K and Velasco-Annis, Clemente and Ouaalam, Abdelhakim and Akhondi-Asl, Alireza and Afacan, Onur and Ortinau, Cynthia M and Clancy, Sean and Limperopoulos, Catherine and Yang, Edward and others, ``A normative spatiotemporal MRI atlas of the fetal brain for automatic segmentation and analysis of early brain growth", \textit{Scientific reports}, vol. 7, no. 1, pp. 476, 2017.
\bibitem{gong2022vqamix} Gong, Haifan and Chen, Guanqi and Mao, Mingzhi and Li, Zhen and Li, Guanbin. Vqamix: Conditional triplet mixup for medical visual question answering, IEEE Transactions on Medical Imaging, vol. 41, no. 11, pp. 3332--3343, 2022. IEEE.
\bibitem{gong2024nnmamba} Gong, Haifan and Kang, Luoyao and Wang, Yitao and Wan, Xiang and Li, Haofeng, ``nnmamba: 3d biomedical image segmentation, classification and landmark detection with state space model", \textit{arXiv preprint arXiv:2402.03526}, vol. , no. , pp. , 2024.
\bibitem{gousias2012magnetic} Gousias, Ioannis S and Edwards, A David and Rutherford, Mary A and Counsell, Serena J and Hajnal, Jo V and Rueckert, Daniel and Hammers, Alexander, ``Magnetic resonance imaging of the newborn brain: manual segmentation of labelled atlases in term-born and preterm infants", \textit{Neuroimage}, vol. 62, no. 3, pp. 1499--1509, 2012.
\bibitem{han2021deep} Han, Xiaoting and Qi, Lei and Yu, Qian and Zhou, Ziqi and Zheng, Yefeng and Shi, Yinghuan and Gao, Yang, ``Deep symmetric adaptation network for cross-modality medical image segmentation", \textit{TMI}, vol. 41, no. 1, pp. 121--132, 2021.
\bibitem{he2022masked} He, Kaiming and Chen, Xinlei and Xie, Saining and Li, Yanghao and Doll{\'a}r, Piotr and Girshick, Ross, ``Masked autoencoders are scalable vision learners", in \textit{Proceedings of the IEEE/CVF conference on computer vision and pattern recognition}, , pp. 16000--16009, 2022.
\bibitem{ho2020denoising} Ho, Jonathan and Jain, Ajay and Abbeel, Pieter, ``Denoising diffusion probabilistic models", \textit{NeurIPS}, vol. 33, no. , pp. 6840--6851, 2020.
\bibitem{hoffman2018cycada} Judy Hoffman, Eric Tzeng, Taesung Park, Jun-Yan Zhu, Phillip Isola, Kate Saenko, Alexei Efros, and Trevor Darrell, ``Cycada: Cycle-consistent adversarial domain adaptation,'' in \textit{International Conference on Machine Learning}, pp. 1989--1998, 2018.
\bibitem{huang2022attentive} Huang, Junjia and Li, Haofeng and Li, Guanbin and Wan, Xiang, ``Attentive symmetric autoencoder for brain MRI segmentation", in \textit{International Conference on Medical Image Computing and Computer-Assisted Intervention}, Springer, pp. 203--213, 2022.
\bibitem{huo2018synseg} Huo, Yuankai and Xu, Zhoubing and Moon, Hyeonsoo and Bao, Shunxing and Assad, Albert and Moyo, Tamara K and Savona, Michael R and Abramson, Richard G and Landman, Bennett A, ``Synseg-net: Synthetic segmentation without target modality ground truth", \textit{TMI}, vol. 38, no. 4, pp. 1016--1025, 2018.
\bibitem{isensee2021nnu} Isensee, Fabian and Jaeger, Paul F and Kohl, Simon AA and Petersen, Jens and Maier-Hein, Klaus H, ``nnU-Net: a self-configuring method for deep learning-based biomedical image segmentation", \textit{Nature methods}, vol. 18, no. 2, pp. 203--211, 2021.
\bibitem{jafari2022lmisa} Mina Jafari, Susan Francis, Jonathan M Garibaldi, and Xin Chen, ``LMISA: A lightweight multi-modality image segmentation network via domain adaptation using gradient magnitude and shape constraint,'' \textit{Medical Image Analysis}, vol. 81, pp. 102536, 2022.
\bibitem{karargyris2023federated} Karargyris, Alexandros and Umeton, Renato and Sheller, Micah J and Aristizabal, Alejandro and George, Johnu and Wuest, Anna and Pati, Sarthak and Kassem, Hasan and Zenk, Maximilian and Baid, Ujjwal and others, ``Federated benchmarking of medical artificial intelligence with MedPerf", \textit{Nature Machine Intelligence}, vol. 5, no. 7, pp. 799--810, 2023.
\bibitem{khader2023denoising} Khader, Firas and M{\"u}ller-Franzes, Gustav and Tayebi Arasteh, Soroosh and Han, Tianyu and Haarburger, Christoph and Schulze-Hagen, Maximilian and Schad, Philipp and Engelhardt, Sandy and Bae{\ss}ler, Bettina and Foersch, Sebastian and others, ``Denoising diffusion probabilistic models for 3D medical image generation", \textit{Scientific Reports}, vol. 13, no. 1, pp. 7303, 2023.
\bibitem{kim2022diffusion} Kim, Boah and Ye, Jong Chul, ``Diffusion deformable model for 4D temporal medical image generation", in \textit{MICCAI}, Springer, pp. 539--548, 2022.
\bibitem{lee2013pseudo} Lee, Dong-Hyun and others, ``Pseudo-label: The simple and efficient semi-supervised learning method for deep neural networks", in \textit{Workshop on challenges in representation learning, ICML}, Atlanta, pp. 896, 2013.
\bibitem{lei2022semi} Lei, Tao and Zhang, Dong and Du, Xiaogang and Wang, Xuan and Wan, Yong and Nandi, Asoke K, ``Semi-supervised medical image segmentation using adversarial consistency learning and dynamic convolution network", \textit{TMI}, vol. , no. , pp. , 2022.
\bibitem{li2020dual} Li, Kang and Wang, Shujun and Yu, Lequan and Heng, Pheng-Ann, ``Dual-teacher: Integrating intra-domain and inter-domain teachers for annotation-efficient cardiac segmentation", in \textit{MICCAI}, Springer, pp. 418--427, 2020.
\bibitem{li2023well} Li, Wenxuan and Yuille, Alan and Zhou, Zongwei, ``How well do supervised models transfer to 3d image segmentation?", in \textit{The Twelfth International Conference on Learning Representations}, , pp. , 2023.
\bibitem{liu2023reducing} S. Liu, S. Yin, L. Qu, and M. Wang, ``Reducing domain gap in frequency and spatial domain for cross-modality domain adaptation on medical image segmentation,'' in \textit{Proceedings of the AAAI Conference on Artificial Intelligence}, vol. 37, no. 2, pp. 1719--1727, 2023.
\bibitem{liu2023structure} S. Liu, S. Yin, L. Qu, M. Wang, and Z. Song, ``A structure-aware framework of unsupervised cross-modality domain adaptation via frequency and spatial knowledge distillation,'' \textit{IEEE Transactions on Medical Imaging}, 2023, IEEE.
\bibitem{makropoulos2018review} Makropoulos, Antonios and Counsell, Serena J and Rueckert, Daniel, ``A review on automatic fetal and neonatal brain MRI segmentation", \textit{NeuroImage}, vol. 170, no. , pp. 231--248, 2018.
\bibitem{moawad2023brain} Moawad, Ahmed W and Janas, Anastasia and Baid, Ujjwal and Ramakrishnan, Divya and Jekel, Leon and Krantchev, Kiril and Moy, Harrison and Saluja, Rachit and Osenberg, Klara and Wilms, Klara and others, ``The brain tumor segmentation (brats-mets) challenge 2023: Brain metastasis segmentation on pre-treatment mri", \textit{ArXiv}, vol. , no. , pp. , 2023.
\bibitem{myronenko20193d} Myronenko, Andriy, ``3D MRI brain tumor segmentation using autoencoder regularization", in \textit{Brainlesion: Glioma, Multiple Sclerosis, Stroke and Traumatic Brain Injuries: 4th International Workshop, BrainLes 2018, Held in Conjunction with MICCAI 2018, Granada, Spain, September 16, 2018, Revised Selected Papers, Part II 4}, Springer, pp. 311--320, 2019.
\bibitem{payette2021automatic} Payette, Kelly and de Dumast, Priscille and Kebiri, Hamza and Ezhov, Ivan and Paetzold, Johannes C and Shit, Suprosanna and Iqbal, Asim and Khan, Romesa and Kottke, Raimund and Grehten, Patrice and others, ``An automatic multi-tissue human fetal brain segmentation benchmark using the fetal tissue annotation dataset", \textit{Scientific Data}, vol. 8, no. 1, pp. 167, 2021.
\bibitem{rastogi2021data} Rastogi, Ananya, ``Data generation for training biomedical models", \textit{Nature Computational Science}, vol. 1, no. 6, pp. 387--387, 2021.
\bibitem{Reinke2023UnderstandingMP} Annika Reinke and Minu Dietlinde Tizabi and Michael Baumgartner and others, ``Understanding metric-related pitfalls in image analysis validation.", \textit{Nature methods}, vol. , no. , pp. , 2023.
\bibitem{sabuncu2010generative} Sabuncu, Mert R and Yeo, BT Thomas and Van Leemput, Koen and Fischl, Bruce and Golland, Polina, ``A generative model for image segmentation based on label fusion", \textit{TMI}, vol. 29, no. 10, pp. 1714--1729, 2010.
\bibitem{sanroma2018learning} Sanroma, Gerard and Benkarim, Oualid M and Piella, Gemma and Camara, Oscar and Wu, Guorong and Shen, Dinggang and Gispert, Juan D and Molinuevo, Jos{\'e} Luis and Ballester, Miguel A Gonz{\'a}lez and Alzheimer’s Disease Neuroimaging Initiative and others, ``Learning non-linear patch embeddings with neural networks for label fusion'', \textit{Medical Image Analysis}, vol.~44, pp.~143--155, 2018.
\bibitem{song2021score} Song, Yang and Sohl-Dickstein, Jascha and Kingma, Diederik P and Kumar, Abhishek and Ermon, Stefano and Poole, Ben, ``Score-based generative modeling through stochastic differential equations", in \textit{ICLR}, , pp. , 2021.
\bibitem{tang2023consistency} Tang, Yongqiang and Wang, Shilei and Qu, Yuxun and Cui, Zhihua and Zhang, Wensheng, ``Consistency and adversarial semi-supervised learning for medical image segmentation", \textit{Computers in Biology and Medicine}, vol. 161, no. , pp. 107018, 2023.
\bibitem{tarvainen2017mean} Tarvainen, Antti and Valpola, Harri, ``Mean teachers are better role models: Weight-averaged consistency targets improve semi-supervised deep learning results", \textit{NeurIPS}, vol. 30, no. , pp. , 2017.
\bibitem{tomar2021self} Tomar, Devavrat and Lortkipanidze, Manana and Vray, Guillaume and Bozorgtabar, Behzad and Thiran, Jean-Philippe, ``Self-attentive spatial adaptive normalization for cross-modality domain adaptation", \textit{TMI}, vol. 40, no. 10, pp. 2926--2938, 2021.
\bibitem{tsai2018learning} Yi-Hsuan Tsai, Wei-Chih Hung, Samuel Schulter, Kihyuk Sohn, Ming-Hsuan Yang, and Manmohan Chandraker, ``Learning to adapt structured output space for semantic segmentation,'' in \textit{Proceedings of the IEEE Conference on Computer Vision and Pattern Recognition}, pp. 7472--7481, 2018.
\bibitem{wang2012multi} Wang, Hongzhi and Suh, Jung W and Das, Sandhitsu R and Pluta, John B and Craige, Caryne and Yushkevich, Paul A, ``Multi-atlas segmentation with joint label fusion", \textit{IEEE transactions on pattern analysis and machine intelligence}, vol. 35, no. 3, pp. 611--623, 2012.
\bibitem{wang2022semi} Wang, Yuchao and Wang, Haochen and Shen, Yujun and Fei, Jingjing and Li, Wei and Jin, Guoqiang and Wu, Liwei and Zhao, Rui and Le, Xinyi, ``Semi-supervised semantic segmentation using unreliable pseudo-labels", in \textit{Proceedings of the IEEE/CVF conference on computer vision and pattern recognition}, , pp. 4248--4257, 2022.
\bibitem{wang2023dhc} Wang, Haonan and Li, Xiaomeng, ``Dhc: Dual-debiased heterogeneous co-training framework for class-imbalanced semi-supervised medical image segmentation", in \textit{International Conference on Medical Image Computing and Computer-Assisted Intervention}, Springer, pp. 582--591, 2023.
\bibitem{wang2024towards} Wang, Haonan and Li, Xiaomeng, ``Towards generic semi-supervised framework for volumetric medical image segmentation", \textit{Advances in Neural Information Processing Systems}, vol. 36, no. , pp. , 2024.
\bibitem{wu2021age} Wu, Jiangjie and Sun, Taotao and Yu, Boliang and Li, Zhenghao and Wu, Qing and Wang, Yutong and Qian, Zhaoxia and Zhang, Yuyao and Jiang, Ling and Wei, Hongjiang, ``Age-specific structural fetal brain atlases construction and cortical development quantification for Chinese population", \textit{Neuroimage}, vol. 241, no. , pp. 118412, 2021.
\bibitem{wu2022exploring} Y. Wu, Z. Wu, Q. Wu, Z. Ge, and J. Cai, ``Exploring smoothness and class-separation for semi-supervised medical image segmentation,'' in \textit{MICCAI}, pp. 34–43, 2022.
\bibitem{wu2023medsegdiff} Wu, Junde and Fu, Rao and Fang, Huihui and Zhang, Yu and Yang, Yehui and Xiong, Haoyi and Liu, Huiying and Xu, Yanwu, ``Medsegdiff: Medical image segmentation with diffusion probabilistic model", in \textit{Medical Imaging with Deep Learning}, PMLR, pp. 1623--1639, 2024.
\bibitem{wu2024medsegdiff} Wu, Junde and Ji, Wei and Fu, Huazhu and Xu, Min and Jin, Yueming and Xu, Yanwu, ``MedSegDiff-V2: Diffusion-Based Medical Image Segmentation with Transformer", in \textit{Proceedings of the AAAI Conference on Artificial Intelligence}, , pp. 6030--6038, 2024.
\bibitem{wu2024voco} Wu, Linshan and Zhuang, Jiaxin and Chen, Hao, ``VoCo: A Simple-yet-Effective Volume Contrastive Learning Framework for 3D Medical Image Analysis", in \textit{CVPR}, , pp. , 2024.
\bibitem{xie2022unsupervised} Xie, Qingsong and Li, Yuexiang and He, Nanjun and Ning, Munan and Ma, Kai and Wang, Guoxing and Lian, Yong and Zheng, Yefeng, ``Unsupervised Domain Adaptation for Medical Image Segmentation by Disentanglement Learning and Self-Training", \textit{TMI}, vol. , no. , pp. , 2022.
\bibitem{xie2023deep} Xie, Long and Wisse, Laura EM and Wang, Jiancong and Ravikumar, Sadhana and Khandelwal, Pulkit and Glenn, Trevor and Luther, Anica and Lim, Sydney and Wolk, David A and Yushkevich, Paul A, ``Deep label fusion: A generalizable hybrid multi-atlas and deep convolutional neural network for medical image segmentation", \textit{MIA}, vol. 83, no. , pp. 102683, 2023.
\bibitem{xu2019larger} Xu, Ruijia and Li, Guanbin and Yang, Jihan and Lin, Liang, ``Larger norm more transferable: An adaptive feature norm approach for unsupervised domain adaptation", in \textit{Proceedings of the IEEE/CVF international conference on computer vision}, , pp. 1426--1435, 2019.
\bibitem{xu2023real} Liuliu Xu, Haifan Gong, Yun Zhong, Fan Wang, Shouxin Wang, Lu Lu, Jinru Ding, Chen Zhao, Wenchao Tang, and Jie Xu, ``Real-time monitoring of manual acupuncture stimulation parameters based on domain adaptive 3D hand pose estimation,'' \textit{Biomedical Signal Processing and Control}, vol. 83, pp. 104681, 2023.
\bibitem{xu2023asc} Xu, Zihang and Gong, Haifan and Wan, Xiang and Li, Haofeng, ``ASC: Appearance and Structure Consistency for Unsupervised Domain Adaptation in Fetal Brain MRI Segmentation", in \textit{MICCAI}, Springer, pp. 325--335, 2023.
\bibitem{yan2022unsupervised} Yan, Pengxiang and Wu, Ziyi and Liu, Mengmeng and Zeng, Kun and Lin, Liang and Li, Guanbin, ``Unsupervised domain adaptive salient object detection through uncertainty-aware pseudo-label learning", in \textit{AAAI}, , pp. 3000--3008, 2022.
\bibitem{yang2020fda} Yang, Yanchao and Soatto, Stefano, ``Fda: Fourier domain adaptation for semantic segmentation", in \textit{CVPR}, , pp. 4085--4095, 2020.
\bibitem{yang2023diffusion} Yang, Ling and Zhang, Zhilong and Song, Yang and Hong, Shenda and Xu, Runsheng and Zhao, Yue and Zhang, Wentao and Cui, Bin and Yang, Ming-Hsuan, ``Diffusion models: A comprehensive survey of methods and applications", \textit{ACM Computing Surveys}, vol. 56, no. 4, pp. 1--39, 2023.
\bibitem{yao2022enhancing} Yao, Huifeng and Hu, Xiaowei and Li, Xiaomeng, ``Enhancing pseudo label quality for semi-supervised domain-generalized medical image segmentation", in \textit{Proceedings of the AAAI conference on artificial intelligence}, , pp. 3099--3107, 2022.
\bibitem{yu2023diffusion} Yu, Xinyi and Li, Guanbin and Lou, Wei and Liu, Siqi and Wan, Xiang and Chen, Yan and Li, Haofeng, ``Diffusion-based data augmentation for nuclei image segmentation", in \textit{MICCAI}, Springer, pp. 592--602, 2023.
\bibitem{yun2019cutmix} Yun, Sangdoo and Han, Dongyoon and Oh, Seong Joon and Chun, Sanghyuk and Choe, Junsuk and Yoo, Youngjoon, ``Cutmix: Regularization strategy to train strong classifiers with localizable features", in \textit{ICCV}, , pp. 6023--6032, 2019.
\bibitem{zhang2018mixup} H. Zhang, M. Cissé, Y. N. Dauphin, and D. Lopez-Paz, ``mixup: Beyond Empirical Risk Minimization,'' in \textit{6th International Conference on Learning Representations, ICLR 2018}, Vancouver, BC, Canada, April 30 - May 3, 2018, Conference Track Proceedings, OpenReview.net, 2018.
\bibitem{zhang2024mapseg} Zhang, Xuzhe and Wu, Yuhao and Angelini, Elsa and Li, Ang and Guo, Jia and Rasmussen, Jerod M and O'Connor, Thomas G and Wadhwa, Pathik D and Jackowski, Andrea Parolin and Li, Hai and others, ``MAPSeg: Unified Unsupervised Domain Adaptation for Heterogeneous Medical Image Segmentation Based on 3D Masked Autoencoding and Pseudo-Labeling'', in \textit{Proceedings of the IEEE/CVF Conference on Computer Vision and Pattern Recognition}, 2024, pp.~5851--5862.
\bibitem{zhao2021mt} Zhao, Ziyuan and Xu, Kaixin and Li, Shumeng and Zeng, Zeng and Guan, Cuntai, ``Mt-uda: Towards unsupervised cross-modality medical image segmentation with limited source labels", in \textit{MICCAI}, Springer, pp. 293--303, 2021.
\bibitem{zhao2022cross} Zhao, Xinkai and Fang, Chaowei and Fan, De-Jun and Lin, Xutao and Gao, Feng and Li, Guanbin, ``Cross-level contrastive learning and consistency constraint for semi-supervised medical image segmentation", in \textit{2022 IEEE 19th International Symposium on Biomedical Imaging (ISBI)}, IEEE, pp. 1--5, 2022.
\bibitem{zhao2022semi} Zhao, Xinkai and Wu, Zhenhua and Tan, Shuangyi and Fan, De-Jun and Li, Zhen and Wan, Xiang and Li, Guanbin, ``Semi-supervised spatial temporal attention network for video polyp segmentation", in \textit{International Conference on Medical Image Computing and Computer-Assisted Intervention}, Springer, pp. 456--466, 2022.
\bibitem{zhao2023masked} Zhao, Xinkai and Hayashi, Yuichiro and Oda, Masahiro and Kitasaka, Takayuki and Mori, Kensaku, ``Masked Frequency Consistency for Domain-Adaptive Semantic Segmentation of Laparoscopic Images", in \textit{International Conference on Medical Image Computing and Computer-Assisted Intervention}, Springer, pp. 663--673, 2023.
\bibitem{zhou2023nnformer} Zhou, Hong-Yu and Guo, Jiansen and Zhang, Yinghao and Han, Xiaoguang and Yu, Lequan and Wang, Liansheng and Yu, Yizhou, ``nnformer: Volumetric medical image segmentation via a 3d transformer", \textit{IEEE Transactions on Image Processing}, vol. , no. , pp. , 2023.
\bibitem{zhu2017unpaired} Jun-Yan Zhu, Taesung Park, Phillip Isola, and Alexei A. Efros, ``Unpaired image-to-image translation using cycle-consistent adversarial networks,'' in \textit{Proceedings of the IEEE International Conference on Computer Vision}, pp. 2223--2232, 2017.
\bibitem{zhuang2016multi} Xiahai Zhuang and Juan Shen, ``Multi-scale patch and multi-modality atlases for whole heart segmentation of MRI,'' \textit{Medical Image Analysis}, vol. 31, pp. 77--87, 2016.
\bibitem{zhuang2019evaluation} Zhuang, Xiahai and Li, Lei and Payer, Christian and {\v{S}}tern, Darko and Urschler, Martin and Heinrich, Mattias P and Oster, Julien and Wang, Chunliang and Smedby, {\"O}rjan and Bian, Cheng and others, ``Evaluation of algorithms for multi-modality whole heart segmentation: an open-access grand challenge'', \textit{Medical Image Analysis}, vol.~58, pp.~101537, 2019.
\bibitem{zou2020unsupervised} Danbing Zou, Qikui Zhu, and Pingkun Yan, ``Unsupervised domain adaptation with dual-scheme fusion network for medical image segmentation,'' in \textit{IJCAI}, pp. 3291--3298, 2020.
\bibitem{myronenko20193d} Andriy Myronenko, ``3D MRI brain tumor segmentation using autoencoder regularization,'' in \textit{Brainlesion: Glioma, Multiple Sclerosis, Stroke and Traumatic Brain Injuries: 4th International Workshop, BrainLes 2018, Held in Conjunction with MICCAI 2018, Granada, Spain, September 16, 2018, Revised Selected Papers, Part II 4}, pp. 311--320, 2019.
\bibitem{payette2023fetal} Kelly Payette, Hongwei Bran Li, Priscille de Dumast, Roxane Licandro, Hui Ji, Md Mahfuzur Rahman Siddiquee, Daguang Xu, Andriy Myronenko, Hao Liu, Yuchen Pei, and others, ``Fetal brain tissue annotation and segmentation challenge results,'' \textit{Medical Image Analysis}, vol. 88, pp. 102833, 2023.
\bibitem{gong2023thyroid} Gong, Haifan and Chen, Jiaxin and Chen, Guanqi and Li, Haofeng and Li, Guanbin and Chen, Fei. Thyroid region prior guided attention for ultrasound segmentation of thyroid nodules, Computers in biology and medicine, vol. 155, pp. 106389, 2023. Elsevier.
\bibitem{gong2024intensity} Gong, Haifan and Huang, Wenhao and Zhang, Huan and Wang, Yu and Wan, Xiang and Shen, Hong and Li, Guanbin and Li, Haofeng. Intensity Confusion Matters: An Intensity-Distance Guided Loss for Bronchus Segmentation, ICME, 2024.
\bibitem{wang2021annotation} Wang, Shanshan and Li, Cheng and Wang, Rongpin and Liu, Zaiyi and Wang, Meiyun and Tan, Hongna and Wu, Yaping and Liu, Xinfeng and Sun, Hui and Yang, Rui and others. Annotation-efficient deep learning for automatic medical image segmentation, Nature communications, vol. 12, no. 1, pp. 5915, 2021. Nature Publishing Group UK London.
\bibitem{ktena2024generative} Ktena, Ira and Wiles, Olivia and Albuquerque, Isabela and Rebuffi, Sylvestre-Alvise and Tanno, Ryutaro and Roy, Abhijit Guha and Azizi, Shekoofeh and Belgrave, Danielle and Kohli, Pushmeet and Cemgil, Taylan and others. Generative models improve fairness of medical classifiers under distribution shifts, Nature Medicine, pp. 1--8, 2024. Nature Publishing Group US New York.
\bibitem{gao2023synthetic} Gao, Cong and Killeen, Benjamin D and Hu, Yicheng and Grupp, Robert B and Taylor, Russell H and Armand, Mehran and Unberath, Mathias. Synthetic data accelerates the development of generalizable learning-based algorithms for X-ray image analysis, Nature Machine Intelligence, vol. 5, no. 3, pp. 294--308, 2023. Nature Publishing Group UK London.
\bibitem{guan2021domain} Guan, Hao and Liu, Mingxia. Domain adaptation for medical image analysis: a survey, IEEE Transactions on Biomedical Engineering, vol. 69, no. 3, pp. 1173--1185, 2021. IEEE.



\end{thebibliography}
\clearpage
\bibliographystyle{alphabetic}

\end{document}